\documentclass[10pt,twocolumn,letterpaper]{article}

\usepackage{cvpr}
\usepackage{times}
\usepackage{epsfig}
\usepackage{graphicx}
\usepackage{amsmath}
\usepackage{amssymb}

\usepackage{comment}

\usepackage[pagebackref=true,breaklinks=true,letterpaper=true,colorlinks,bookmarks=false]{hyperref}

\cvprfinalcopy 

\def\cvprPaperID{1856} 

\newcommand{\rei}[1]{\textcolor[rgb]{0.0,0,0.0}{#1}}
\newcommand{\moriwaki}[1]{\textcolor[rgb]{0.0,0,0.0}{#1}}

\newcommand{\yoshic}[1]{\textcolor[rgb]{0.0,0.0,0.0}{[Yoshi:#1]}}
\newcommand{\yoshi}[1]{\textcolor[rgb]{0.0,0.0,0.0}{#1}}
\newcommand{\yoshia}[1]{\textcolor[rgb]{0.0,0.0,0.0}{#1}} 
\newcommand{\yoshiaa}[1]{\textcolor[rgb]{0.0,0.0,0.0}{#1}} 
\newcommand{\yoshiaaa}[1]{\textcolor[rgb]{0.0,0.0,0.0}{#1}} 

\newcommand{\shao}[1]{\textcolor[rgb]{0.0,0.0,0.0}{#1}}

\newcommand{\rec}{$\mathcal{L}_{Rec}$}
\newcommand{\vgg}{$\mathcal{L}_{VGG}$}
\newcommand{\gan}{$\mathcal{L}_{GAN}$}

\renewcommand{\paragraph}[1]{\vspace{1mm} \noindent  {\bf {#1}}　\hspace{1mm}}

\newcommand{\bm}{\boldsymbol}

\ifcvprfinal\fi
\begin{document}

\title{Hybrid Loss for Learning Single-Image-based HDR Reconstruction}

\author{Kenta Moriwaki$^1$ \\  \\ \vspace{-4mm} \\ $^{1}$The University of Tokyo\\
{\tt\small }
\and
Ryota Yoshihashi$^1$ \\ \hspace{-50mm}Shaodi You$^{2,3}$ \\  \vspace{-4mm} \\ $^{2}$Data61-CSIRO \\
\and
Rei Kawakami$^{1}$ \\ \hspace{-50mm}Takeshi Naemura$^{1}$ \\ \vspace{-4mm} \\$^{3}$Australian National University
\and 
{ \vspace{-80mm} \tt\footnotesize \{moriwaki,yoshi,rei,naemura\}@nae-lab.org, Shaodi.You@data61.csiro.au }
}

\makeatletter
\let\@oldmaketitle\@maketitle
\renewcommand{\@maketitle}{\@oldmaketitle
\vspace{-11mm}
\begin{center}
\includegraphics[width=0.91\textwidth]{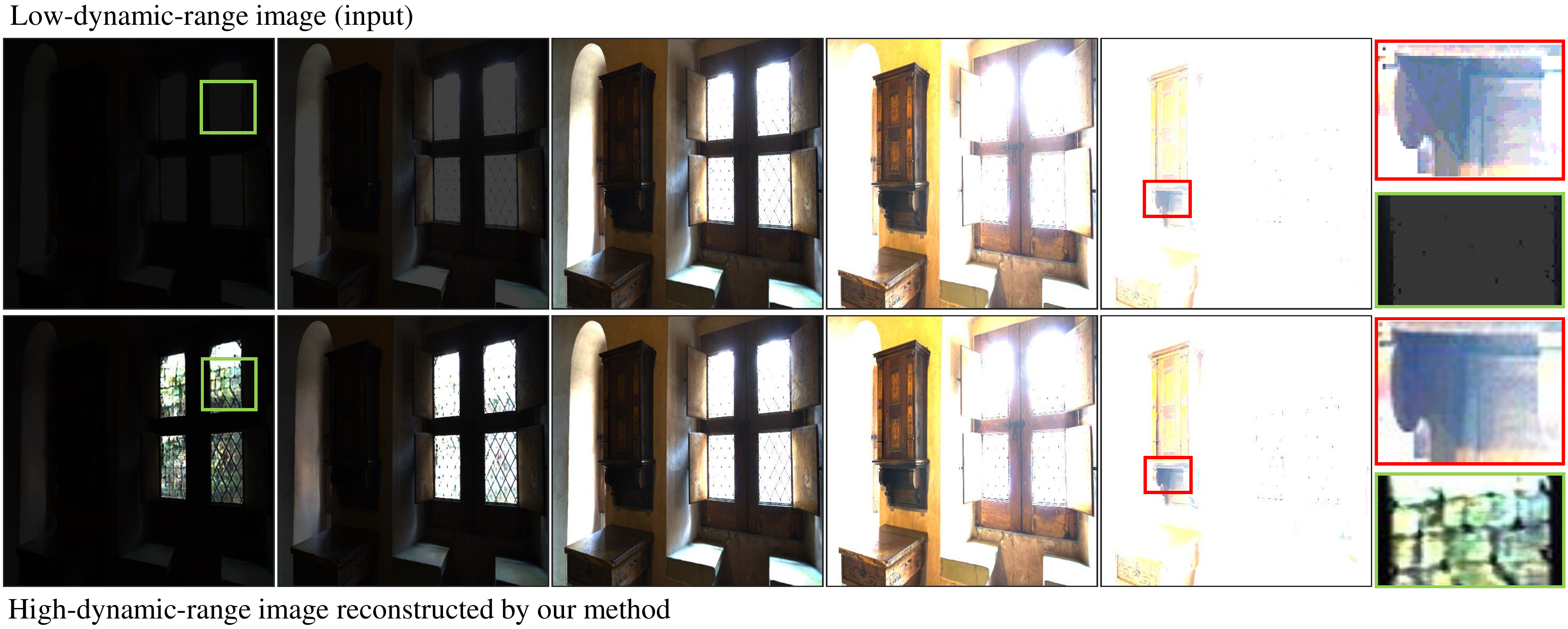}
\end{center}
\vspace{-3mm}
{
\small
Figure 1: 
The top row shows multiple exposure levels of an LDR image, the original exposure level of which is the third one from left.
Given the original LDR image as an input, our method
reconstructs an HDR image, as shown in the bottom row with the multiple exposure levels. 
No structures in the over-exposed region (green) are visible in the LDR image and the under-exposed region (red) is grossly quantized. Our method can inpaint these lost structures plausibly and recover intensity gradients in the under-exposed region.
     }
\bigskip}
\makeatother
\setcounter{figure}{1}

\maketitle

\begin{abstract}
\vspace{-5.5mm}
This paper tackles high-dynamic-range (HDR) image reconstruction given only a single low-dynamic-range (LDR) image as input. While the existing methods focus on minimizing the mean-squared-error (MSE) between the target and reconstructed images, 
we minimize a hybrid loss that consists of perceptual and adversarial losses in addition to HDR-reconstruction loss. The reconstruction loss \shao{instead of} MSE is more suitable for HDR \shao{since} 
it puts more weight on both over- and under- exposed areas. It makes the reconstruction faithful to the input. Perceptual loss
enables the networks to utilize knowledge about objects and image structure for recovering the intensity gradients of saturated and grossly quantized areas. Adversarial loss helps to select the most plausible appearance from multiple solutions. The hybrid loss that combines all \shao{the three} losses is calculated in logarithmic space of image intensity so that the outputs retain a large dynamic range and meanwhile the learning becomes tractable.
Comparative experiments conducted with other state-of-the-art methods demonstrated that our method produces a leap in image quality.
\end{abstract}

\section{Introduction}
\vspace{-2mm}
High-dynamic-range (HDR) imaging is capable of expressing a wide range of light intensities. It can avoid over- and under- exposures, and can express image brightnesses beyond the quantization resolution of the sensor. It enhances the viewing experience when images are shown on HDR displays. Moreover, it is used in image-based rendering for accurate simulation of environmental lighting, and in artistic image editing, owing to its rich representation capability. For an overview of the technique, see~\cite{reinhard2010high,Banterle2017,mantiuk2015}. 
To obtain an HDR image, we currently need either an expensive HDR camera or multiple shots from a low-dynamic-range (LDR) camera with different exposures~\cite{hdr1997}. However, if an HDR image could be reconstructed from a single LDR image, billions of photographs shot in LDR could be utilized for HDR applications. 
%
%

To enable single-image HDR reconstruction, conventional inverse-tone-mapping (iTM) methods focus on saturated areas and extrapolate the light intensity from the surrounding regions on  the basis of local heuristics. As the heuristics are not universal, they often fail to recover over- and under- exposed areas or introduce unnatural artifacts. More recently, deep-learning-based methods have been applied to single-image-based HDR reconstruction~\cite{endoSA2017,HDRsingle2017,Marnerides_2018_EG,Lee2018DeepCH,Yang_2018_CVPR}. However, the images produced by the existing methods 
still suffer from artifacts and insufficient contrast. This is partly because they only focus on minimizing the mean-squared error (MSE) between the reconstructed and target images. 
%

Single-image-based HDR reconstruction is a highly ill-posed problem that prevents MSE-based learning from producing satisfactory images. An ideal HDR reconstruction must have the following properties: 1) Inpainting of over-exposed areas: clipped values due to saturation must be extrapolated for higher intensity, but 
since the information is lost from those regions, \shao{how to extrapolate them is often rather ambiguous}. 2) Recovering the intensity gradients of under-exposed areas: due to quantization, gradations are lost in very dark regions, which causes unnatural artifacts. 3) Visual fidelity: after inpainting and restoration, the reconstructed HDR image must appear natural to the human eye. Blur or artifacts harm the visual impression even if they are negligible when measured by MSE. 4) Semantic consistency: inpainted or restored HDR images must be plausible with the context of the scene and its objects. For example, in a sunset image, saturated regions in the sky may have to be re-colorized in reddish orange. 
Given such four properties, multiple solutions may exist, and as MSE-based solutions tend to be the average of a number of possible solutions, they may often be blurred.

This hybrid nature of HDR reconstruction makes it difficult to design one good loss function. To overcome the problem,
we introduce a {\it hybrid loss} that consists of HDR-reconstruction loss, adversarial loss~\cite{GAN2014}, and perceptual loss~\cite{Gatys2015TextureSU,Johnson2016Perceptual,SISR2017}.
{HDR-reconstruction loss} \yoshiaaa{is our novel loss that} minimizes per-pixel difference between images, and is useful for image reconstruction. \shao{This} loss is designed to put more weight on over- and under- exposed areas so that it \yoshiaaa{achieves} the properties of 1) and 2). 
{Adversarial loss} is utilized for better image quality, addressing the property of 3). In contrast to the other losses which are fixed during optimization, the loss from the framework of generative adversarial networks (GAN)~\cite{GAN2014} can be formulated as a two-player game where the generator and the discriminator are updated competitively and whose solution is the Nash equilibrium between the players. A GAN can select a solution from among the set of possible ones~\cite{DBLP:journals/corr/Goodfellow17}. Therefore, it may find a better solution than those only relying on MSE minimization. 
{Perceptual loss} is a high-level similarity between images, which can be calculated by using the output of the high-level layer in pre-trained convolutional neural networks (CNN) that encode broad knowledge of object appearances. The loss causes the reconstructed image to be more natural and plausible, addressing the property of 4).

All of the losses are \shao{essential} to guide the optimization of our network, and multiple losses can also avoid artifacts that may be produced by a single loss measures. To ensure the output values \shao{that} have a high dynamic range and still keep the learning tractable, all losses are calculated in logarithmic space of image intensity. This is found to be  very effective. 
We also introduce a new evaluation protocol for HDR image reconstruction that considers the intensity gradient recovery in under-exposed regions as equally as that in over-exposed regions. 
The experiments show that the proposed method makes a leap in the resulted image quality compared to the state-of-the-art methods.

%
%
%



%

The contributions of the paper are summarized as follows. 
First, we propose a hybrid loss consisting of reconstruction loss, adversarial loss, and perceptual loss. The reconstruction loss is devised so that the intensity gradients in saturated and dark regions can be reconstructed. All losses are defined in logarithmic space of image intensity, which is crucial to make the learning tractable. To our knowledge, this is the first study that introduces a perceptual loss for HDR reconstruction. Second, the results of the proposed method are a leap in quality in comparison with the state of the art. Third, we introduced a new evaluation protocol that takes the under-exposed regions into account as well as the over-exposed regions. Finally, we collected a new HDR dataset that are publicly available on the web, and will release the URL list of the images. The code and trained model will be published upon acceptance of this paper.


\section{Related work}
\vspace{-2mm}

\paragraph{Single-image HDR reconstruction}
Conventionally, HDR reconstruction has been performed by non-learning-based brightness enhancement through filtering or light-source detection. For example, bilateral filters applied to $x$-$y$-range three-dimensional grids work as brightness enhancement functions~\cite{kovaleski2009high,KovaleskiOliveira2014}. 
However, non-learning-based approaches cannot estimate physically accurate amounts of light due to the lack of knowledge about real HDR images; thus, the quality of the estimated HDR images is limited.

A few studies have used deep learning in HDR reconstruction from a single LDR image. Such methods can be categorized into multi-step and single-step methods. The multi-step methods generate bracketed images with multiple exposures and then merge them. The single-step methods generate an HDR image directly in one step.

An example of a multi-step methods is Deep Reverse Tone Mapping (DrTMO)~\cite{endoSA2017}, which generates 
multiple images with different exposures using an encoder-decoder network~\cite{autoencoder2006,VincentPLarochelleH2008}. To train the network, LDR images are simulated using various camera curves~\cite{GrossbergCVPR2003} from an HDR image dataset and input.
ChainHDRI~\cite{Lee2018DeepCH} and RecursiveHDRI~\cite{Lee2018ECCV} are similar to DrTMO~\cite{endoSA2017}, the difference being that they recurrently generate higher or lower exposure images from images generated in the previous time steps. However, such recurrent methods need multiple forward computations in one HDR generation; in contrast, ours can generate HDR images in one forward pass.

With the growing popularity of end-to-end learning, single-step networks
that directly estimate the desired HDR images may be preferable to multi-step methods.
HDR-CNN~~\cite{HDRsingle2017} and Deep Reciprocating HDR~\cite{Yang_2018_CVPR} share the same encoder-decoder structure that directly generates an HDR image from an LDR image. While the architecture itself is similar to UNet for segmentation~\cite{Unet2015}, they train networks to recover from over-/under-exposures of moderate extent that are artificially added to the training LDR images.
ExpandNet~\cite{Marnerides_2018_EG} \shao{has} a three-branch architecture designed for single-step HDR image generation, \shao{and the branches} are for global, semi-local, and local feature extraction. In contrast, we show that the simple encoder-decoder architecture performs well with our augmented loss functions.

\paragraph{GANs}
The essential difficulty with single-image HDR reconstruction is in the restoration of over- or under-exposed regions, where structures in the original scenes are totally lost or heavily corrupted. Even the deep-learning methods discussed above 
suffer from imperfect restoration and unnatural artifacts.
For this reason, GANs~\cite{GAN2014}, which \shao{have} successfully restored and inpainted natural appearing images~\cite{pathakCVPR16context, Yang2017CVPR}, are considered promising. 
If the reconstructive error (\shao{for example, the} MSE between the outputs and the training images) is the only loss function, the restored images are easily blurred. A GAN can mitigate such artifacts and recover more detailed texture. 

GANs have already been used for HDR image generation, by Lee \etal~\cite{Lee2018ECCV} and Ning \etal~\cite{GAN_HDR_2018}. By introducing GAN, \yoshiaa{the restoration quality is further improved
than the simple encoder-decoder networks}. 
We found that GAN combined with reconstructive error still generates blur or unnatural artifacts. In this paper, by further introducing perceptual loss and reconstruction loss optimized for HDR, the image quality can be improved.

\paragraph{Deep image processing}
\yoshia{Apart from HDR reconstruction, we can see wider variety of deep-learning methods for image processing within LDR images, which are still useful as references. 
For example, convolutional GANs \shao{well-performed} in superresolution~\cite{SISR2017}, denoising~\cite{chen2018image}, or inpaiting~\cite{pathakCVPR16context}.
Other than GANs, there are some promising approaches such as multiscale~\cite{Yang2017CVPR,lu2018deep}, perceptual losses~\cite{Johnson2016Perceptual}, attention~\cite{Qian_2018_CVPR}, or reinforcement learning~\cite{yu2018crafting,Furuta2019AAAI}. 
While their insights are useful also for our task, such methods for LDR images are not directly applicable to HDR images.}


\begin{figure}[t]
	\begin{center}
      \includegraphics[width=0.97\hsize]{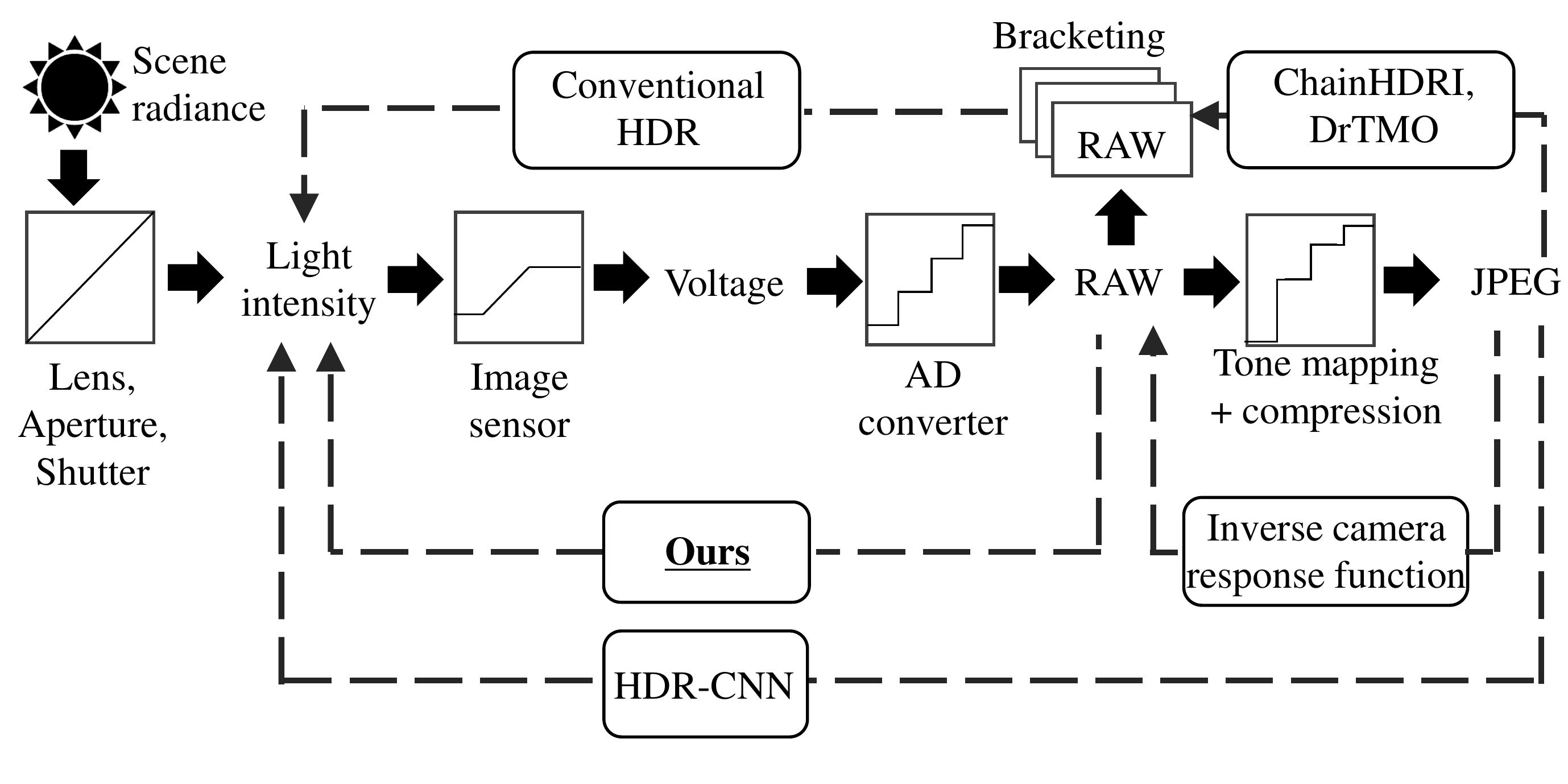}
	\end{center}
    \vspace{-2mm}
    \caption{The pipelines of image formation in cameras and HDR reconstruction methods. Ours estimates HDR light intensity from linearized (raw) images, while other deep-learning-based methods use images after tone mapping and compression. \yoshiaaa{Our method can accurately estimate the light intensity using measurement-based linearization.}}
    \vspace{-4mm}
    \label{fig:pipeline}
\end{figure}

\section{Method}
\vspace{-1mm}
\subsection{Problem statement}
\vspace{-1mm}
\yoshia{Single-image-based HDR reconstruction can be defined as a task to estimate physical light intensity from single LDR images. 
Since the imaging pipeline in cameras is a lossy process, estimating physical light intensity from RAW or JPEG images is an ill-posed problem.
The imaging pipeline~\cite{kim2012new}, as shown in Fig~\ref{fig:pipeline}, consists of the followings: First, a lens gather\shao{s} light rays and forms an image on a sensor.
The amount of light that reaches the sensor is controlled by an aperture and shutter speed, which decides \shao{the} exposure value of the image. The sensor outputs voltages corresponding to the amount of light, but too large voltages are cropped due to saturation. The voltages from the sensor are digitized by an AD converter, and in this part the small voltage values are quantized, leading to the lost tones in under-exposed regions. 
The images after AD conversion are called RAW images, and \shao{they} are further \shao{tone-mapped} and compressed into JPEG images. From such images, single-image-based HDR reconstruction methods need \shao{to} estimate the original light intensity.}

\yoshiaa{Given the pipeline of image formation, there is a degree of freedom in from which stage a method reconstruct HDR images.}
\yoshia{The most conventional way to reconstruct HDR images from LDR images is exposure bracketing~\cite{hdr1997}, which is to capture a single scene by multiple LDR images with various exposure values and merge them later into an HDR image. 
In single-image-based HDR reconstruction, 
\shao{most of the} learning-based HDR-reconstruction methods use JPEG images after tone mapping~\cite{HDRsingle2017,endoSA2017}.
While this is useful for applying to daily JPEG images,
it may make the reconstruction more difficult.
The tone mapping makes \shao{the} nonlinearity between light intensity and pixel values larger. In addition, the mapping functions differ by cameras, \shao{which increases} the uncertainty of reconstruction.  
In contrast, we estimate HDR images from raw images, which preserve linear relationship to light intensity within non-saturating regions. 
This does not reduce the applicability of our method, since raw images can be easily recovered from JPEG images when the camera response function is known, and even when it is unknown, a number of methods are available to estimate the inverse camera response function from images~\cite{grossberg2003determining,takamatsu2008estimating}.}

\begin{figure}[t]
	\begin{center}
  ％\includegraphics[width=0.9\hsize]{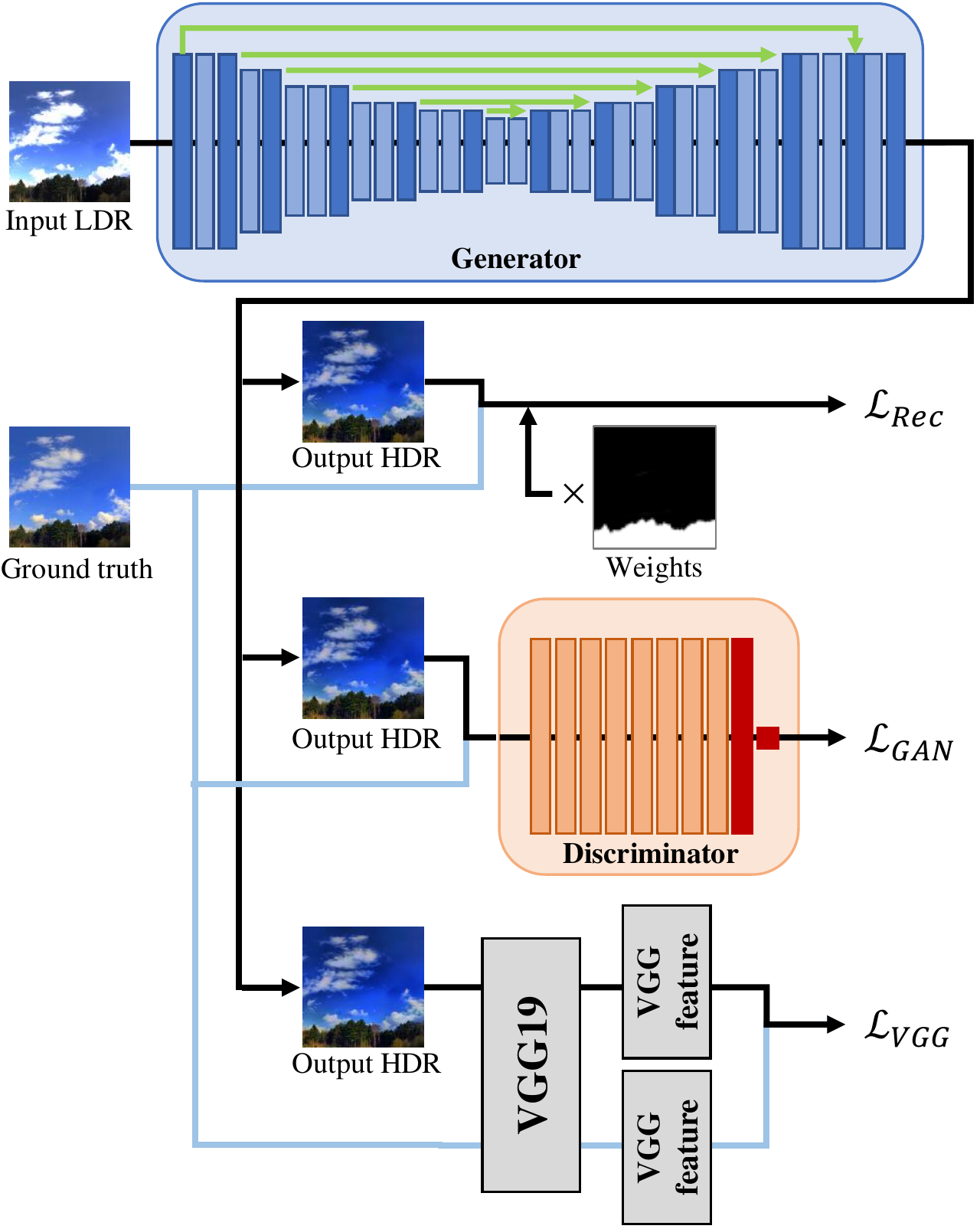}
	\end{center}
    \vspace{-2mm}
    \caption{\yoshi{The overview of our method. Our generator is an encoder-decoder network \rei{with skip connections}. In addition to the adversarial loss \rei{$\mathcal{L}_{GAN}$} from 
\rei{the} discriminator, we also incorporate \rei{HDR-reconstruction loss $\mathcal{L}_{Rec}$} 
and 
\rei{perceptual loss $\mathcal{L}_{VGG}$} to improve image quality.
\vspace{-3mm}
}}
        \label{fig:system_overview}
\end{figure}
\vspace{-1mm}

\subsection{The hybrid loss}
\vspace{-1mm}
The outline of the method is shown in Fig.\,\ref{fig:system_overview}.
\rei{The generator is 
an encoder-decoder network with skip connections.}
The input for the generator is a color LDR image with 8 bits per channel, and the output 
is an image of the float data type with 32 bits per channel.\footnote{Note that our method outputs much more bits compared to that of the HDR image format.} 
%
The detail on how to synthesize an input LDR image is described in Sec.~\ref{subsec:datasets}.

To train the generator, we use a hybrid loss, combining 
\rei{HDR-reconstruction loss $\mathcal{L}_{Rec}$, adversarial loss $\mathcal{L}_{GAN}$, and perceptual loss $\mathcal{L}_{VGG}$.}
The hybrid 
loss of the generator $\mathcal{L}_G$ can be written as follows\shao{:}
\begin{equation} \label{eqn:hybrid}
		\mathcal{L}_G = \mathcal{L}_{Rec}  + \alpha \mathcal{L}
_{GAN} + \beta \mathcal{L}_{VGG},
\end{equation}
where \rei{$\alpha$ and $\beta$ are weights to balance the losses.}

\paragraph{HDR-reconstruction loss}
\yoshia{Our $\mathcal{L}_{Rec}$, the \rei{reconstruction loss}, penalizes the per-pixel $\ell^2$ distance of intensities between the reconstructed and the ground-truth HDR images.} \yoshia{A problem in defining $\mathcal{L}_{Rec}$ \rei{w.r.t} HDR images is the wide range of values; naive loss functions, such as the mean-squared error, may depend too much on the high-luminance regions, and errors in the lower range will be negligible.}
To avoid this, we define the loss in the logarithmic domain of pixel intensity. We also introduce weights to put more attention on over- and under- exposed regions.
Thus, the $\mathcal{L}_{Rec}$ can \rei{be} expressed as follows:
\begin{equation}\label{eq:recloss}
\mathcal{L}_{Rec}(\hat{y},y) = \frac{1}{N} \sum_{i=1}^{N}
w_i 
{|
     	\log{(\hat{y}_i}) - \log{({y}_i}}) 
 |^2,
\end{equation}
\rei{where $\hat{y}$ is the reconstructed HDR image by the generator, $y$ is the original HDR image in the training set (ground truth), $w_i$ corresponds to the pixel-wise weights, $i$ is for each pixel,  and $N$ is the number of pixels.} 
\yoshia{For simplicity, Eq.~\ref{eq:recloss} shows the HDR-reconstruction loss in the per-image form, but the loss is averaged over the mini batch during training.}

Since over-exposed regions can naturally have large intensity difference, we introduce weights for emphasizing under-exposed regions. 
Thus, $ w_i $ is defined as
\begin{equation}
\yoshia{
w_i =  1 + \gamma\max\left(0, 1 - \frac{1}{\tau}\min_c(x_{i,c})\right),}
\label{eq:underexposed}
\end{equation}
where $x_{i,c}$ is the LDR image normalized to $[0, 1]$, and $c$ is for color channel. $\tau$ is the threshold, and \yoshia{ $\gamma$ is \rei{a weight to enhance} the loss of under-exposed regions.} \rei{In this paper, the threshold $\tau$ is set to $0.05$.}

\paragraph{Adversarial loss}
The adversarial loss $\mathcal{L}_{GAN}$, as in a GAN, is introduced so that the generator can \rei{deceive} the discriminator. 
The loss is useful for making the generated images \shao{close} to the \yoshi{distribution of the original dataset.} It is expressed as follows:
\begin{equation}
	\mathcal{L}_{GAN} = \sum^{M}_{j=1} -\log D( \log(\hat{y}_j)), \label{eq:LGAN}
\end{equation}
where \rei{$D(\hat{y})$ is the probability of classifying whether $\hat{y}$ is real or fake, and $M$ is the number of training images in the mini batch. Here, we also use logarithm of $\hat{y}$ for calculating the loss, to keep the high dynamic range of the images and still make the learning tractable. The loss is smaller when $\hat{y}$ is closer to the original input, such that the discriminator $D$ considers it is real.} 

The discriminator $D$ is \rei{composed of eight convolutional layers and two fully connected layers.} In the training, the input for the discriminator is \rei{either of a pair, one of which}
is $\hat{y}$ and the other is $y$.
\rei{The loss function $\mathcal{L}_D$} for training the discriminator can be written as follows~\cite{GAN2014}:
\begin{equation}
	\begin{split}
		&\mathcal{L}_D = - \frac{1}{M}\sum^{M}_{j=1} \biggl(\log \bigl(1-D(\, \log(\hat{y}_j)\,)  \bigr) + \log \bigl( D(\log(y_j)) \bigr) \biggr).        
	\end{split}
    \label{eq:discriminator}
\end{equation}
The logarithm of $\hat{y}$ and $y$ is introduced \rei{for the same reason as in Eq.~\ref{eq:LGAN}.}
The two losses $\mathcal{L}_G$ and $\mathcal{L}_D$ are used for updating the weights of the generator and \shao{the} discriminator, respectively.

\paragraph{Perceptual loss}
\yoshia{The perceptual loss enhances perceptual similarity for human eyes and mitigate artifacts in the output image by utilizing pre-trained image-classification networks.} 
We adopt VGG19~\cite{VGG19} for this purpose following~\cite{Bruna2015SuperResolutionWD,SISR2017,Johnson2016Perceptual}.
\yoshia{VGG19 is pre-trained in ILSVRC2012, which is an LDR-image-classification dataset, and it is not directly applicable to HDR images due to the difference of domains. However, we found that the perceptual loss is still useful by applying logarithmic  transformation on input HDR images.}
Specifically, we denote the output of the \shao{$l$-th} pooling layer in VGG19 as $\bm{\phi}_{l}$ , and $\mathcal{L}_{VGG}$ is defined as follows:
\begin{equation}
\mathcal{L}_{VGG}(\hat{y},y) = \frac{1}{N_{l}}\sum_{k=1}^{N_{l}}
    |\bm{\phi}_{l}(\log(\hat{y}))_{k} - \bm{\phi}_{l}(\log(y))_{k}|^2,
\end{equation}
where $N_{l}$ represents the total number of pixels in the feature space, \yoshia{and $\bm{\phi}_{l}(\cdot)_k$ represents the feature vector at the pixel $k$}. In this paper, we used $l=5$ in the same manner as in \cite{SISR2017}. 
\yoshia{The perceptual loss is also averaged over the mini batch during training.}

\section{Experiments}
\yoshia{We \rei{conducted} experiments on publicly available HDR image sets to compare the reconstructed HDR quality of our method and existing ones. \rei{Furthermore}, we show extensive visualization of HDR-reconstruction results 
and analysis.} 

\paragraph{Datasets}
\label{subsec:datasets}
\yoshia{We used a part of the dataset used in HDR-CNN~\cite{HDRsingle2017} and images crawled from the web newly by us. 
The motivation is that approximately half of the training images used in~\cite{HDRsingle2017} are private data \shao{of} the author \shao{which are not} publicly available.
Due to the lack of public large-scale HDR image set sufficient to train deep networks, prior studies
partially used private HDR image sets~\cite{HDRsingle2017,Lee2018DeepCH,Lee2018ECCV}, which may be a problem in reproduction.
In contrast, all of our training data is publicly available on the web.}
\moriwaki{Our dataset consists of images of indoor and outdoor \yoshia{scenes}. We used 999 HDR images and 61 HDR videos for training.
We will release the URL list of the images, although \rei{distribution of the original images is not allowed due to the copyright.}}

\begin{table*}[t]
\caption{\moriwaki{Comparison of the ground truth and HDR images by the proposed and other state-of-the-art methods. The input images have brightness coressponding to $2^{-6}$. (See Fig.\ref{fig:data_example} for the corresponding range.)}}
\vspace{1mm}
\label{table:result_eye}
\centering
\scalebox{0.85}{
  \begin{tabular}{|c|c|c|c|c|c|c|c|c|c|c|} \hline
  & \multicolumn{4}{|c|}{Reinhard's TMO} &  \multicolumn{4}{|c|}{Kim and Kautz's TMO} &  \multicolumn{2}{|c|}{VDP quality}\\ 
         & \multicolumn{2}{|c}{PSNR(dB)} &  \multicolumn{2}{c|}{SSIM} &  \multicolumn{2}{|c}{PSNR(dB)} &  \multicolumn{2}{c|}{SSIM} & \multicolumn{2}{|c|}{score}\\ \cline{2-11}
         
         & $m$ & $\sigma$ & $m$ & $\sigma$ & $m$ & $\sigma$ & $m$ & $\sigma$ & $m$ & $\sigma$  \\ \hline \hline
         
    Proposed                         & \bf{31.53} & 5.60 & \bf{0.948} & 0.028 & \bf{29.71} & 2.74 & \bf{0.931} & 0.034 & \bf{51.30} & 4.79 \\ \hline
    HDR-CNN~\cite{HDRsingle2017}         & 18.33 　　　& 1.27 & 0.791      & 0.063 & 21.68      & 2.28 & 0.846      & 0.061 & 51.09
& 4.57 \\ \hline \hline  
    DrTMO~\cite{endoSA2017}            & 23.70 　　　& 7.26 & 0.846      & 0.165 & 21.97      & 7.72 & 0.819      & 0.184 & 43.59
& 3.70 \\ \hline    
    ExpandNet~\cite{Marnerides_2018_EG}    & 18.60 　　　& 3.28 & 0.729      & 0.129 & 17.35      & 2.43 & 0.721      & 0.093 & 46.92
& 6.16 \\ \hline    
    Huo \etal.~\cite{HuoYDB14}              & 14.97 　　　& 0.90 & 0.665      & 0.072 & 15.82      & 1.62 & 0.716      & 0.069 & 39.77
& 3.87 \\ \hline    
    KOEO \cite{KovaleskiOliveira2014} & 16.75 　　　& 1.71 & 0.706      & 0.066 & 16.96      & 2.32 & 0.737      & 0.069 & 39.00
& 3.10 \\ \hline \hline
   RecursiveHDRI~\cite{Lee2018ECCV}          & 26.71 　　　& 2.78 & \slash     & \slash & 22.31      & 3.20 & \slash      & \slash & 48.85 & 4.91 \\ \hline
  \end{tabular}}
\end{table*}

\begin{figure}[t]
	\begin{center}
        \includegraphics[width=\hsize]{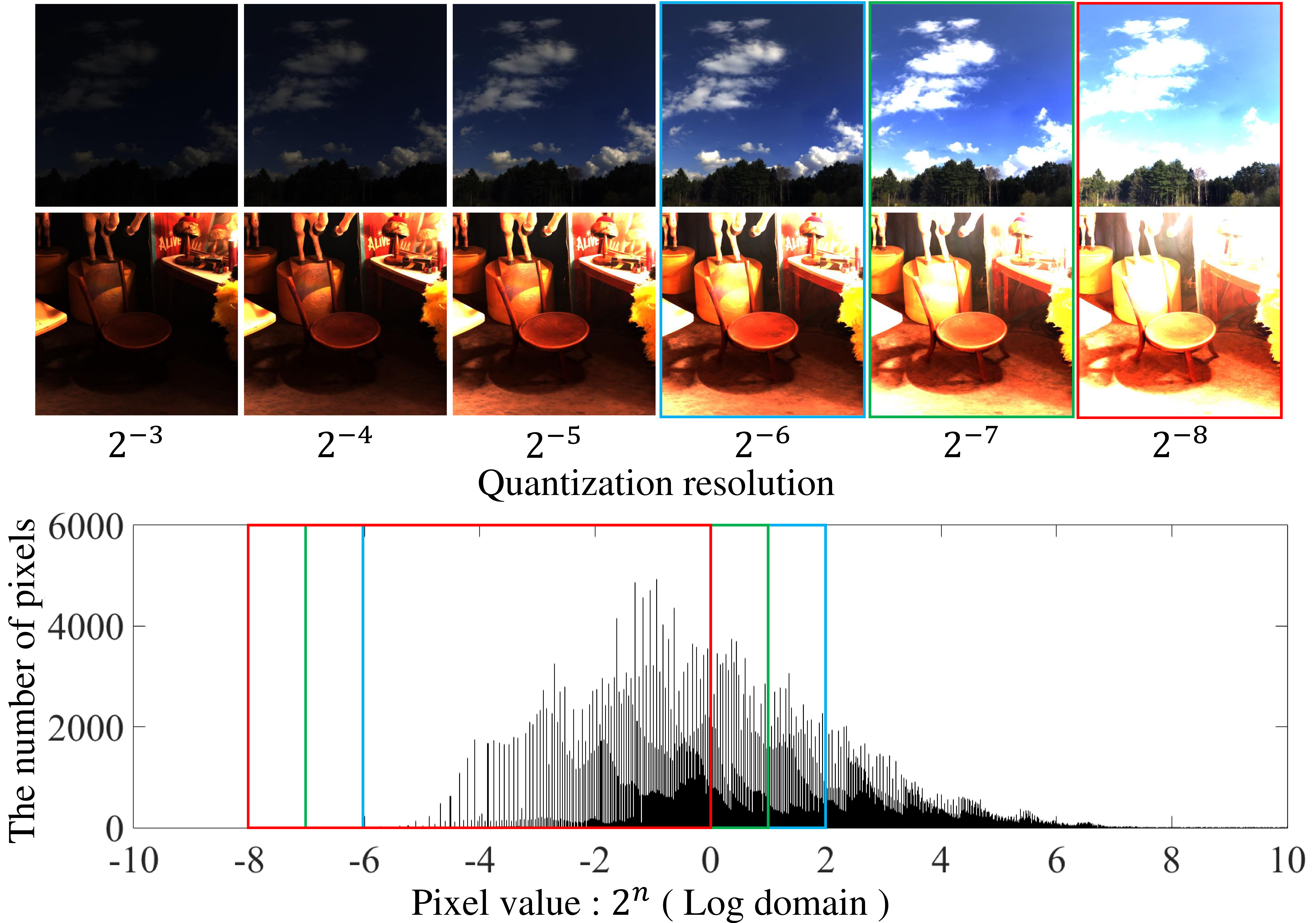}
     \vspace{-6mm}
	\end{center}
    \caption{
    \yoshia{The visualization of decomposed LDR images from an HDR image. Each exposure level is referred by its quantization resolution. The lower figure shows the entire histogram of HDR pixel values and LDRs capture part of it. The red, green, and blue rectangles correspond to the same exposure level in the top row. }
    }
    \vspace{-3mm}
    \label{fig:data_example}
\end{figure}

\yoshia{
To train an HDR reconstruction network,
we need a collection of pairs, one of which is an HDR image as ground truth and the other is an LDR image as the input. Thus, we automatically generated LDR images from the HDR dataset. First, the HDR images were cropped at random positions and resized to $256\times256$. 
Then, each HDR image was normalized by its median of the luminance value.
Finally, we decomposed the HDR image into multiple LDR images with various quantization resolutions, each of which had the range of 8-bit for each channel.  
Figure~\ref{fig:data_example} shows \rei{the visualization of the range of decomposed LDR images.} 
Specifically, we decomposed an HDR image into six 
\rei{exposure levels,} 
and we refer to each of the levels by using their quantization resolution. 
\rei{Among the six decomposed images, images in the five ranges corresponding to the quantization resolution of $2^{-8}$ to $2^{-4}$ are randomly used as inputs in the training. 
The total number of LDR images for training is 127,831. }}

\begin{figure*}[t]
	\begin{center}
        \includegraphics[width=16.5cm]{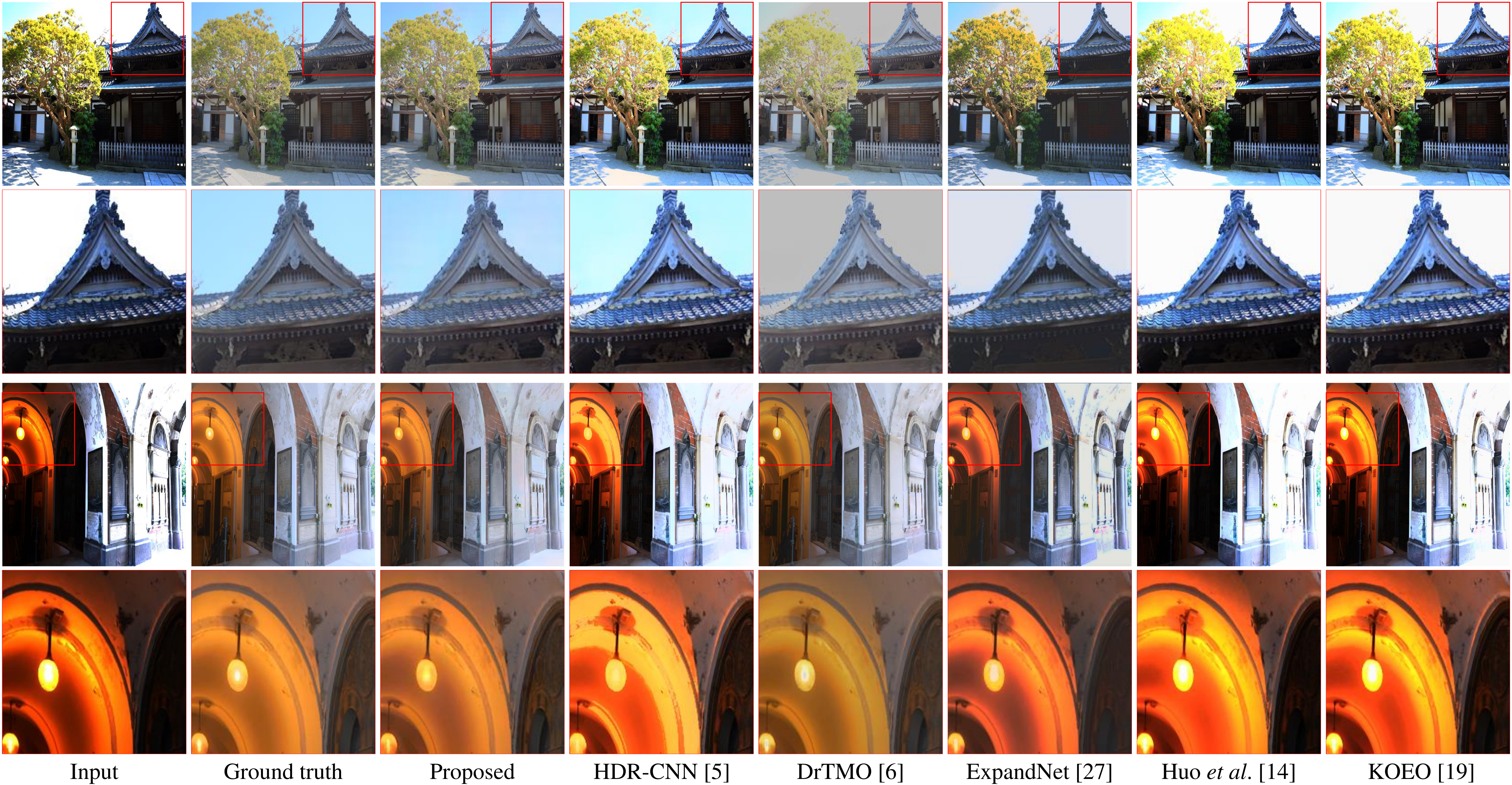}
	\end{center}
    \vspace{-3mm}
    \caption{Comparison \yoshiaaa{between} the ground truth and HDR images reconstructed  by the proposed and other methods. HDR images are tone-mapped using the method of Reinhard \etal.~\cite{Reinhard2002} }
        \label{fig:Result_eye_TMO}
\end{figure*}

\paragraph{Training}
\yoshia{\rei{We followed the GAN training framework~\cite{GAN2014}, where a generator and a discriminator were updated alternatively.
We used our hybrid loss (Eq.~\ref{eqn:hybrid}) to update our generator, and the discriminator loss (Eq.~\ref{eq:discriminator}) to update our discriminator during training. 
The loss minimization was performed with \shao{the} ADAM optimizer~\cite{kingma2014adam}. 
For ADAM's parameter, the initial learning rate was set to $2.0\times10^{-5}$, the batch size was 16, and the total number of epochs was 60. }
For the parameters in the hybrid loss,
$\alpha$ and $\beta$ in Eq.~\ref{eqn:hybrid} were set as follows: 
$\alpha=1.0\times10^{-3}, \beta=5.0 \times10^{-4}$. $\alpha$ was set to align the order of magnitude of $L_{Rec}$ and $L_{GAN}$, and $\beta$ was set so that $L_{VGG}$ is approximately $10$ times larger than the other loss functions.}
\rei{$\gamma$ was set to $5.0$, so that the under-exposed region was enhanced approximately $5$ times.} 

\moriwaki{We \rei{utilized the weights of} the trained model of HDR-CNN~\cite{HDRsingle2017} as the initial values of the generator. \rei{In HDR-CNN, the model was pre-trained with Places database~\cite{Places_Database}, and then trained using the collected HDR-image dataset.}}
\rei{We fine-tuned the model further with our dataset. The network architecture of the discriminator is based on~\cite{SISR2017}, which is composed of 10 layers, 8 of which are convolutional layers and 2 are fully connected layers. It is trained from scratch. }

\begin{table}[t]
\caption{PSNR(dB) compared in the LDR image stacks. (See Fig.~\ref{fig:data_example} for the corresponding range.)}
\vspace{1mm}
\label{table:result_PSNR}
\centering
\scalebox{0.75}{
  \begin{tabular}{|c|cccccc|c|} \hline
    Method & $2^{-3}$ & $2^{-4}$ & $2^{-5}$ & $2^{-6}$ & $2^{-7}$ & $2^{-8}$ & Mean \\ \hline \hline
    Proposed & \bf{34.57} & \bf{30.66} & \bf{27.67} & \bf{27.82} & \bf{32.44} & \bf{32.16} & \bf{30.88}  \\ \hline
    HDR-CNN~\cite{HDRsingle2017} & 27.66 & 23.20 & 19.44 & 17.67 & 18.90 & 16.39 & 20.54 \\ \hline \hline  
    DrTMO~\cite{endoSA2017} & 28.56 & 24.05 & 20.26 & 18.97 & 23.19 & 26.59 & 23.60 \\ \hline    
    ExpandNet~\cite{Marnerides_2018_EG} & 25.44 & 20.89 & 17.22 & 16.15 & 19.95 & 18.84 & 19.74 \\ \hline    
    Huo \etal.~\cite{HuoYDB14} & 19.09 & 14.13 & 11.35 & 12.74 & 16.48 & 14.50 & 14.71 \\ \hline    
    KOEO~\cite{KovaleskiOliveira2014} & 17.13 & 12.26 & 11.26 & 13.50 & 16.97 & 15.21 & 14.38 \\ \hline 
  \end{tabular}}
  \vspace{-3mm} 
\end{table}

\begin{table}[t]
\caption{SSIM compared in the LDR image stack.}
\vspace{1mm}
\label{table:result_SSIM}
\centering
\scalebox{0.75}{
  \begin{tabular}{|c|cccccc|c|} \hline
    Method & $2^{-3}$ & $2^{-4}$ & $2^{-5}$ & $2^{-6}$ & $2^{-7}$ & $2^{-8}$ & Mean \\ \hline \hline
    Proposed & \bf{0.937} & \bf{0.931} & \bf{0.931} & \bf{0.951} & \bf{0.978} & \bf{0.964} & \bf{0.948}  \\ \hline
    HDR-CNN~\cite{HDRsingle2017} & 0.833 & 0.784 & 0.752 & 0.753 & 0.782 & 0.793 & 0.782 \\ \hline \hline  
    DrTMO~\cite{endoSA2017} & 0.859 & 0.828 & 0.812 & 0.843 & 0.906 & 0.953 & 0.866 \\ \hline    
    ExpandNet~\cite{Marnerides_2018_EG} & 0.786 & 0.738 & 0.718 & 0.756 & 0.841 & 0.877 & 0.786 \\ \hline    
    Huo \etal.~\cite{HuoYDB14} & 0.597 & 0.534 & 0.516 & 0.581 & 0.681 & 0.753 & 0.610 \\ \hline    
    KOEO~\cite{KovaleskiOliveira2014} & 0.567 & 0.515 & 0.534 & 0.616 & 0.706 & 0.762 & 0.616 \\ \hline
  \end{tabular}}
  \vspace{-4mm}
\end{table}

\begin{figure*}[t]
	\begin{center}
		\includegraphics[width=16.5cm]{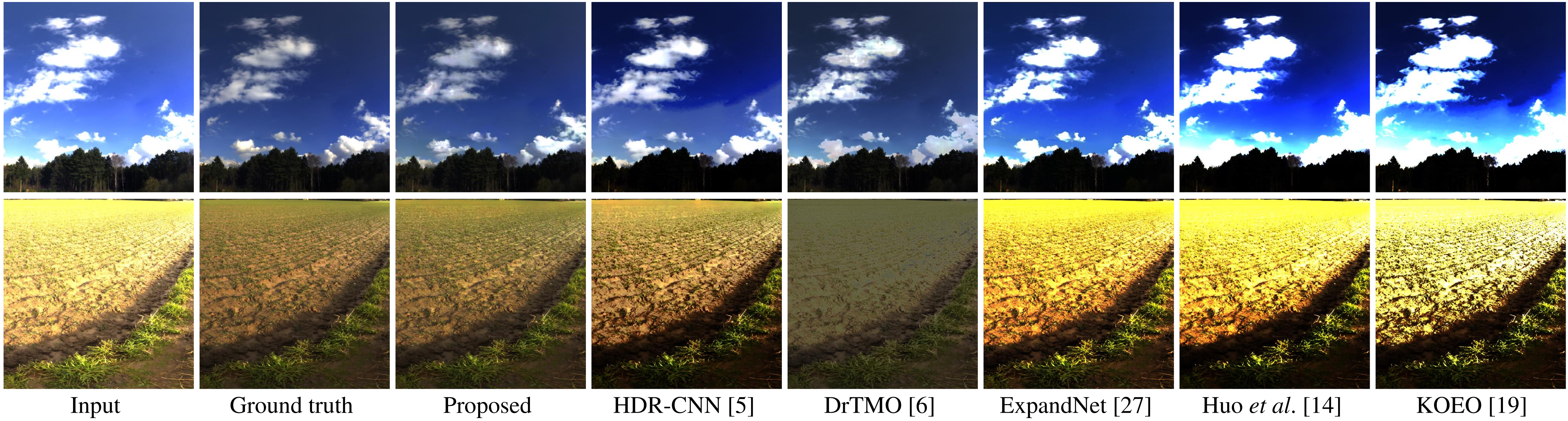}
	\end{center}
    \vspace{-3mm}
    \caption{\moriwaki{Comparison of the ground truth LDR and reconstructed LDR images. The image corresponding to the range of $2^{-7}$ was used for the input, and the estimated HDR image was decomposed into six exposure levels of LDR images and evaluated. The figure shows the ground truth and the results of each method corresponding to the range of $2^{-6}$ in Fig.~\ref{fig:data_example}.}}
        \label{fig:Result1}
\end{figure*}

\paragraph{Comparisons with state-of-the-art methods}
\yoshia{First, we show the quantitative comparisons to existing single-image-based HDR reconstruction methods. Following the latest work~\cite{Lee2018ECCV},
we use HDREye dataset~\cite{Nemoto2015VisualAI} for the test set.
We report PSNR and SSIM between the ground truth and each HDR image inferred by the methods, both of which were tone-mapped by the tone-mapping operator (TMO) of Reinhard \etal~\cite{Reinhard2002} and Kim and Kautz~\cite{Kim2008}. 
Also, we report the metric of HDR-VDP-2~\cite{HDRVDP2_2011}, which is based on the human visual system to evaluate the estimated HDR images. 
The parameters used for HDR-VDP-2 are exactly the same as in~\cite{Lee2018ECCV}: a 24-inch display with a resolution of $1,900\times1,200$, and a view distance of 0.5 m.}

Table~\ref{table:result_eye} and Fig.~\ref{fig:Result_eye_TMO} show the evaluation results. 
\moriwaki{We compared \rei{ours with the iTM methods of Huo \etal~\cite{HuoYDB14} and Kovaleski and Oliveira expansion operator (KOEO)~\cite{KovaleskiOliveira2014}}, in addition to DrTMO~\cite{endoSA2017}, HDR-CNN~\cite{HDRsingle2017} and ExpandNet~\cite{Marnerides_2018_EG}, which are the state-of-the-art methods using deep learning. Table~\ref{table:result_eye} is divided into three blocks. The \rei{top block shares the same training dataset of ours.} The middle block is the results of using \rei{the models that have been released to public, though the training dataset may be partially different from ours (though we believe there should be a lot of overlap)}. The \rei{bottom} block is the results \rei{reported in RecursiveHDRI~\cite{Lee2018ECCV} using exactly the same test set and evaluation protocol, though we do not have access to the network \shao{or} the dataset.} DrTMO~\cite{endoSA2017} \rei{is supposed to use the processed image with camera response function as an input, but we report the results inputting the linear LDR image (the same as others), since the scores are slightly better with the linear inputs.}}

\moriwaki{
\rei{As Table~\ref{table:result_eye} shows}, the proposed method is superior to the other methods in all the metrics. As shown in Fig.~\ref{fig:Result_eye_TMO}, even if qualitatively evaluated, we can see that our results are closer to the ground truth images. \rei{As shown in the images in the top two rows, the saturated regions of the sky area were successfully recovered by the proposed method. As shown in the images in the bottom two rows, the texture inside the light bulb is successfully recovered.}}

\begin{figure*}[t]
	\begin{center}
        \includegraphics[width=16.0cm]{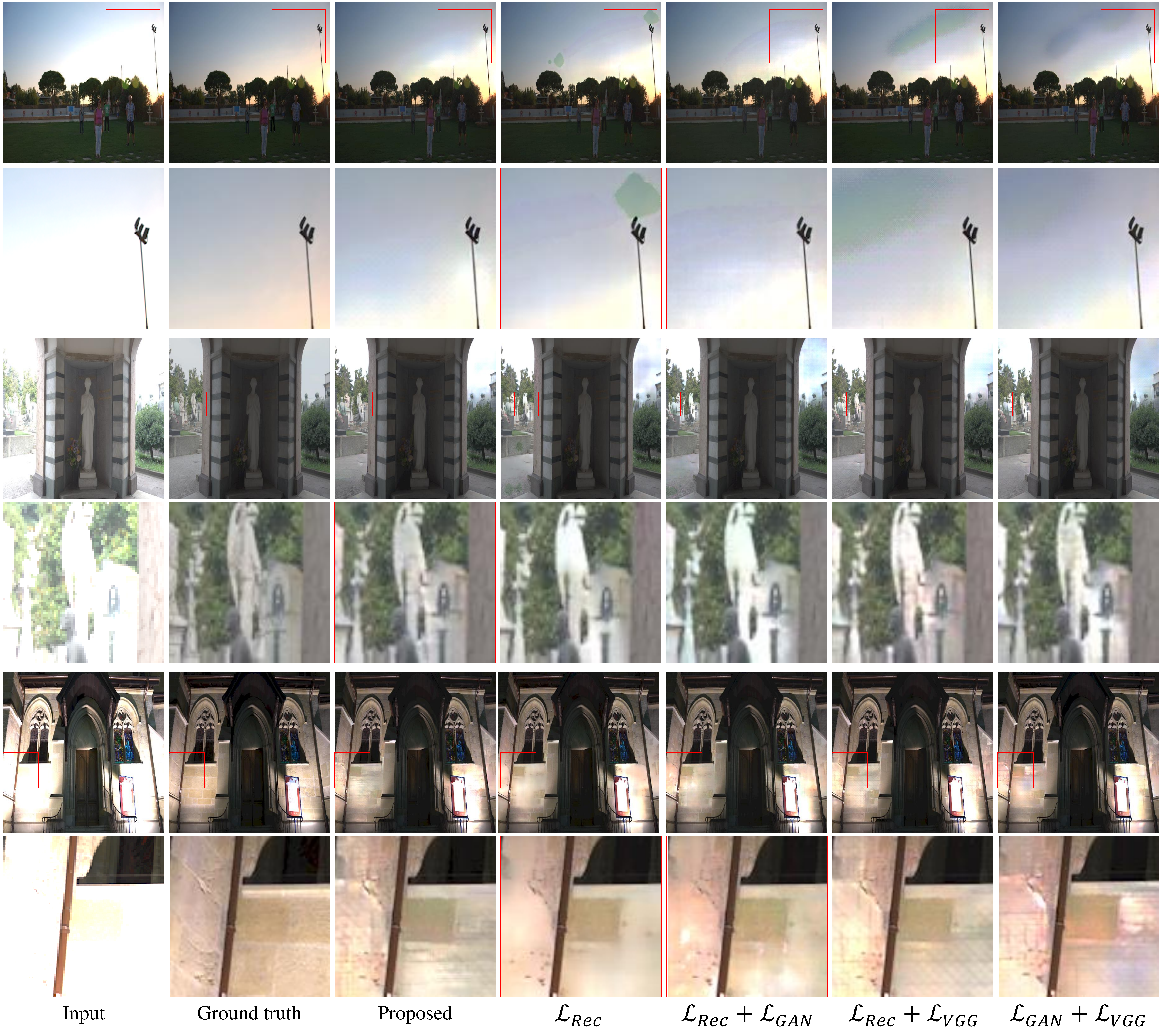}
	\end{center}
    \vspace{-4mm}
    \caption{\moriwaki{Comparison of results using different combinations of the loss functions, \rec, \gan, and \vgg. The results of the proposed method are visually the closest to the ground truth, and have less artifacts.} }
        \label{fig:compare_cost_func}
\end{figure*}

\paragraph{Range-wise evaluation}
\label{subsec:comparison}
\rei{In the evaluation metrics in Table~\ref{table:result_eye}, the errors in the bright regions will be dominant and under-exposed regions will have negligible effect. To visualize the errors at each exposure levels equally, in this paper, we introduce a new evaluation protocol; namely, the errors are evaluated in the decomposed LDR images, ranging from $2^{-3}$ to $2^{-8}$. For this evaluation, our dataset was split into training and testing sets. We used 900 images for training. For testing, we excluded images that may be used for training in HDR-CNN~\cite{HDRsingle2017}, and used the \shao{remainder which contain} 33 images. The input LDR images are in the range of $2^{-7}$.}

\rei{
Tables~\ref{table:result_PSNR} and \ref{table:result_SSIM}\yoshiaaa{,} and Fig.~\ref{fig:Result1} show the results evaluated with the decomposed LDR image sets. Tables~\ref{table:result_PSNR} and \ref{table:result_SSIM} include two blocks as in Table~\ref{table:result_eye}, where the difference between them is the training datasets.
\yoshiaaa{PSNR} and SSIM shown in Tables~\ref{table:result_PSNR} and \ref{table:result_SSIM} respectively show that
the proposed method is superior to other methods in all the ranges.
As shown in Fig.\ref{fig:Result1}, 
iTMs fail to recover over-exposed areas, and \shao{lead} to unnatural boundaries in the near-saturation areas. The proposed method can recover the contrast accurately, and the results are visually closer to the ground truth image. More results can be found in the supplementary material.}
\begin{figure}[htbp]
  \begin{center}
  \includegraphics[width=\columnwidth]{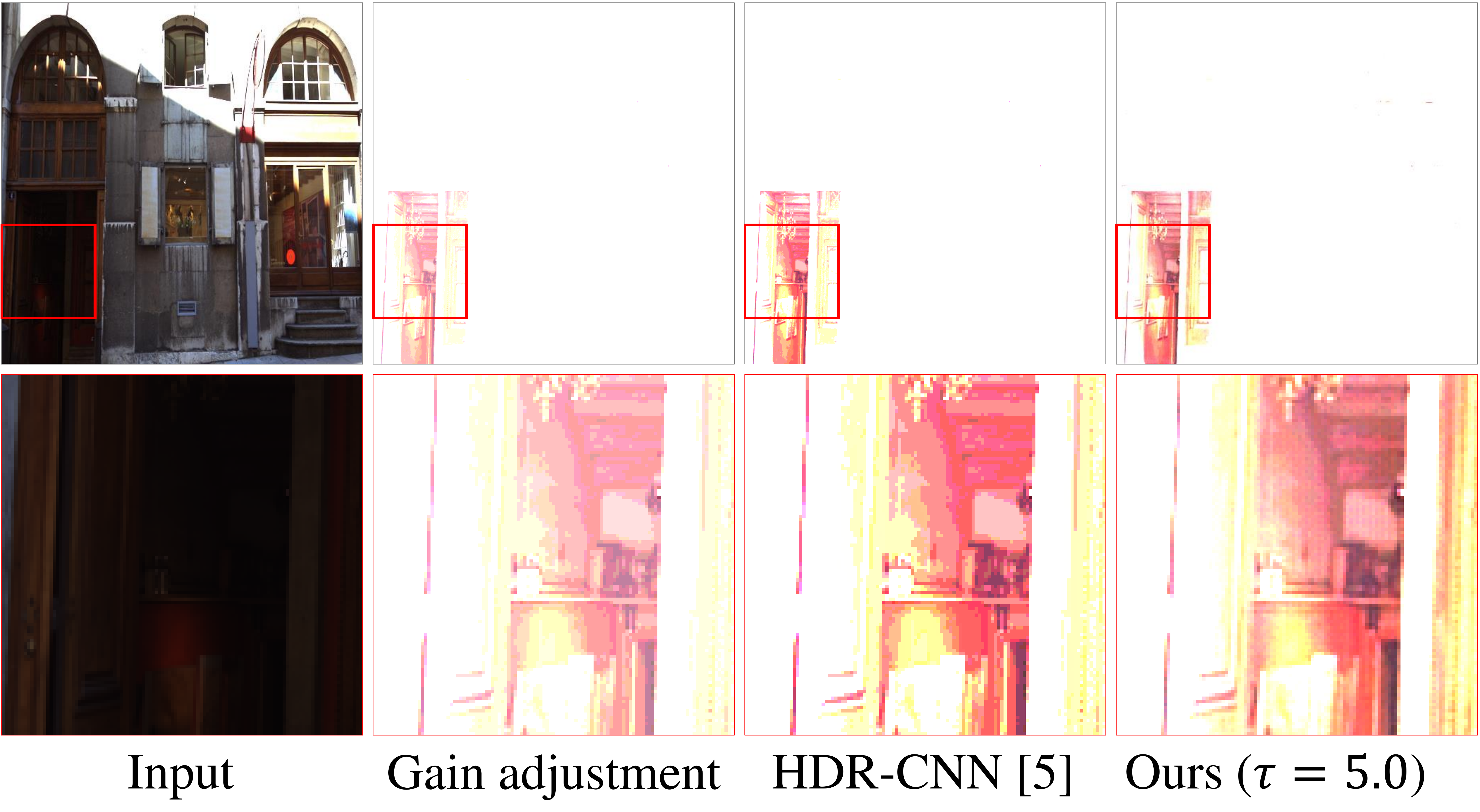}
  \begin{center}
  \vspace{-3mm}
  \caption{Zoomed views of under-exposed region. The intensity gradients are recovered successfully in those grossly quantized areas with our method. (Best viewed in color with zoom in.)}
  \vspace{-12mm} 
   \label{fig:under-exposure}
  \end{center}
  \end{center}
  
\end{figure}
\paragraph{Ablation study}
\moriwaki{
\rei{We compare \yoshiaaa{our full hybrid loss} with its ablations. We use the HDREye dataset and the range-wise evaluation protocol for this study. The input images are either in the range of $2^{-5}$ or $2^{-6}$.
Table~\ref{table:cost_function} and Fig.~\ref{fig:compare_cost_func} show the results of the combinations of the three losses, \rec, \gan, and \vgg, used for the generator.}
\shao{Although in Table~\ref{table:cost_function} the PSNR and SSIM are slightly degraded by introducing \gan\ or \vgg, as shown in Fig.~\ref{fig:compare_cost_func}, \rei{
result images such as those using \gan$+$\vgg\ are visually more plausible, while results of
\rec\ are smoothed and texture-less.}}
By using the hybrid loss, the reconstructed image is visually plausible and also faithful to the input image. In addition, each of the networks, \shao{i.e.,} the generator, the discriminator, and the VGG19, has unique artifacts because of aliasing of using the spatial re-sampling. By combining all of them, such patterns can be reduced effectively.
}

\begin{table}[t]
\caption{\moriwaki{Ablation study of the loss functions. In terms of PSNR and SSIM, reconstruction loss \rec\ is the most effective. However, $\mathcal{L}_{GAN}$ and $\mathcal{L}_{VGG}$ produce visually plausible results as shown in Fig.~\ref{fig:compare_cost_func}. } }
\vspace{1mm}
\label{table:cost_function}
\centering
\scalebox{0.9}{
  \begin{tabular}{c|cccc|c} 
        &  &  &  &  & Ours \\ \hline
    $\mathcal{L}_{Rec}$ & \checkmark & \checkmark & \checkmark &  & \checkmark     \\ 
    $\mathcal{L}_{GAN}$ &  & \checkmark &  & \checkmark & \checkmark   \\   
    $\mathcal{L}_{VGG}$ &  &  & \checkmark & \checkmark & \checkmark   \\ \hline
    PSNR & \bf{31.37} & 30.83 & 30.69 & 29.54 & 30.75   \\ \hline 
    SSIM & \bf{0.957} & 0.951 & 0.952 & 0.945 & 0.953   \\  
  \end{tabular}}
  \vspace{-3mm} 
\end{table}

\vspace{-1mm} 
\paragraph{Restoration of under-exposed areas}
\moriwaki{Most existing methods focus on the saturated region for intensity recovery. We show that our method is effective for restoring the dark region as shown in Fig.~\ref{fig:under-exposure}. Information in the under-exposed region is grossly quantized as can be seen in the image with \shao{the} adjusted gain. The boundaries of the staircase and the texture of the walls are restored with the weights in Eq.~\ref{eq:underexposed}.}

\vspace{-1mm} 
\section{Conclusion}
\vspace{-1mm} 
We \shao{presented} a method to reconstruct HDR images directly from a single LDR image by designing a hybrid loss incorporating HDR reconstruction loss, adversarial loss, and perceptual loss, \shao{which are all} calculated using logarithmic space of image intensity. The method produces superior results compared with existing methods, and successfully recovers both over- and under- exposed regions. 

The limitation is that when saturated areas are too large, the generator struggles to inpaint the regions. Deepening the network, collecting larger datasets, and conducting user study are our future work.

\section{Acknowledgement}
This work was also supported by JSPS KAKENHI Grant Number JP18K11348,
and Grant-in-Aid for JSPS Fellows JP16J04552. The authors
would like to thank Dr. Ari Hautasaari for his advice to improve the manuscript and Mr. Shun Iwasawa in DWANGO Co., Ltd. for his advice on this research.

{\small
\bibliographystyle{ieee}
\bibliography{mybib}
}
 
\section{List of the HDR Dataset}
Table~\ref{table:hdr_list} shows the list of URLs that provide the datasets used in our paper for training and testing. Copyrights of the images are owned by the photographers and the authors of the websites. HDRIhaven  and HDReye are only used for testing in our experiments. Please note that the two links to EMPA and Pouli\_hdrldr are no longer available as of submission, and for those data, we recommend readers contacting the copyright holders or us directly.
\begin{table*}[h]
\caption{URL list of datasets}
\vspace{1mm}
\label{table:hdr_list}
\centering
\scalebox{0.75}{
  \begin{tabular}{|c|l|l|c|} \hline
     & Name & URL & Size  \\ \hline \hline
Image & HDRIhaven & \small \url{https://hdrihaven.com/hdris/} & 248  \\ \cline{2-4}   
& Funt & \small \url{http://www.cs.sfu.ca/~colour/data/funt_hdr/\#DATA} & 107  \\ \cline{2-4}   
   & Fairchild & \small \url{http://rit-mcsl.org/fairchild//HDRPS/HDRthumbs.html} & 104  \\ \cline{2-4} 
   & Stanford & \small \href{http://scarlet.stanford.edu/~brian/hdr/hdr.html}{http://scarlet.stanford.edu/$\sim$brian/hdr/hdr.html} & 88  \\ \cline{2-4}    
     & HDRMAPS & \small \url{http://hdrmaps.com/freebies} & 71  \\ \cline{2-4} 
& HDR-dataset & \small \url{http://www.hdrlabs.com/sibl/archive.html} & 64  \\ \cline{2-4}   
     & HDReye & \small \url{https://mmspg.epfl.ch/hdr-eye} & 46  \\ \cline{2-4} 
     & Ward & \small \url{http://www.anyhere.com/gward/hdrenc/pages/originals.html} & 33  \\ \cline{2-4}   
     & Freeskies & \small \url{https://joost3d.com/hdris/} & 23  \\ \cline{2-4}       
     & Noemotion & \small \url{http://noemotionhdrs.net/hdrother.html} & 21  \\ \cline{2-4}
     
     & BOCO & \small \url{http://bocostudio.com/boco-pano/} & 15  \\ \cline{2-4}
     & Dutch$\textunderscore$360 & \small \url{https://www.dutch360hdr.com/shop/product-category/free-360-hdri/} & 14  \\ \cline{2-4}    
     & Openfootage & \small \url{http://www.openfootage.net/category/high-dynamic-range-panorama/hdris-with-a-much-higher-dynamic-range/} & 14  \\ \cline{2-4}
     & HDRI$\textunderscore$hub & \small \url{https://www.hdri-hub.com/hdrishop/freesamples/freehdri/item/323-hdr-city-road-night-lights-free} & 11  \\ \cline{2-4}    

     & Viz people & \small \url{https://www.viz-people.com/portfolio/free-hdri-maps/} & 10 \\ \cline{2-4}    
     & pfstools & \small \url{http://pfstools.sourceforge.net/hdr_gallery.html} & 9  \\ \cline{2-4}  
     & Giantcow & \small \url{http://giantcowfilms.com/2015/11/23/hdr-morning-sun-winter/} & 4  \\ \cline{2-4}    
     & HDRishop & \small \url{https://www.hdrishop.com/collections/free-hdris/products/free-bathroom-hdri} & 3  \\ \cline{2-4}
     & Dylan sisson & \small \url{http://www.dylansisson.com/project/panoramas/} & 2  \\ \cline{2-4}    
     & EMPA & \small http://www.empamedia.ethz.ch/hdrdatabase/index.php & 33  \\ \cline{2-4}
     & Pouli$\textunderscore$hdrldr & \small Statistical Regularities in Low and High Dynamic Range Images by Pouli et al. [2010] & 327   \\ \hline \hline 
Video  & Stuttgart & \small \url{https://hdr-2014.hdm-stuttgart.de/} & 33  \\ \cline{2-4}
& DML-HDR & \small \url{http://dml.ece.ubc.ca/data/DML-HDR/} & 10  \\ \cline{2-4}  
     & LiU HDRV & \small \url{http://hdrv.org/Resources.php} & 10  \\ \cline{2-4}  
     & Boltard & \small \url{https://people.irisa.fr/Ronan.Boitard/} & 7   \\ \cline{2-4}   
     & MPI & \small \url{http://resources.mpi-inf.mpg.de/hdr/video/} & 1 \\ \hline     
  \end{tabular}}
\end{table*}
\section{Additional Experimental Results}
\subsection{Range-wise evaluation}
\begin{figure*}[t]
	\begin{center}
		\includegraphics[width=16.5cm]{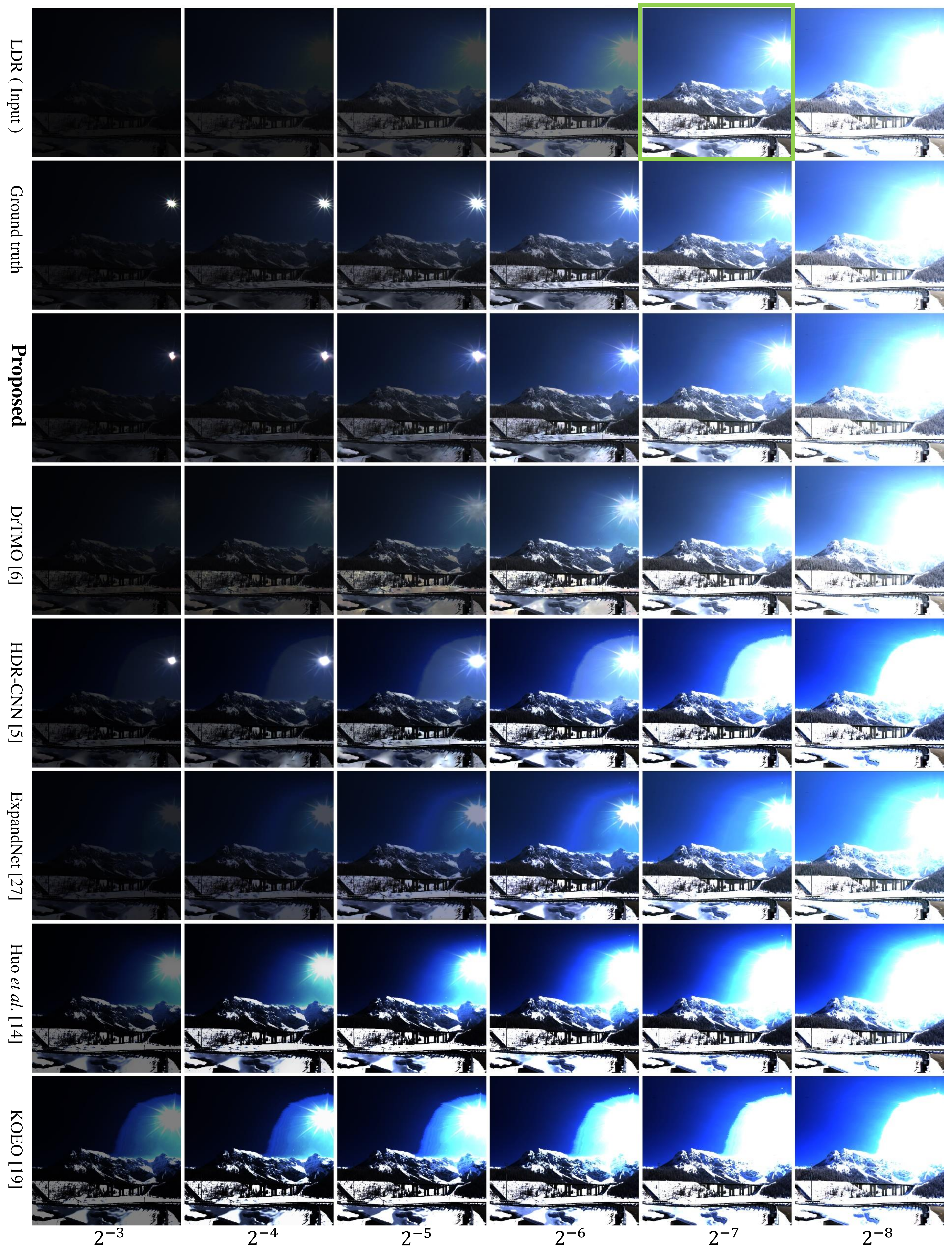}
	\end{center}
    \vspace{-3mm}
    \caption{The top row shows multiple exposure levels of an LDR image (the original LDR is highlighted with a green rectangle), where each column corresponds to the levels from $2^{-8}$ to $2^{-3}$. The second row shows the ground truth, and the rest are the HDR images reconstructed by each method, given the LDR image as an input. The light intensity of the sun is partly recovered by the proposed method.}
        \label{fig:Result_o1}
\end{figure*}

\begin{figure*}[t]
	\begin{center}
		\includegraphics[width=16.5cm]{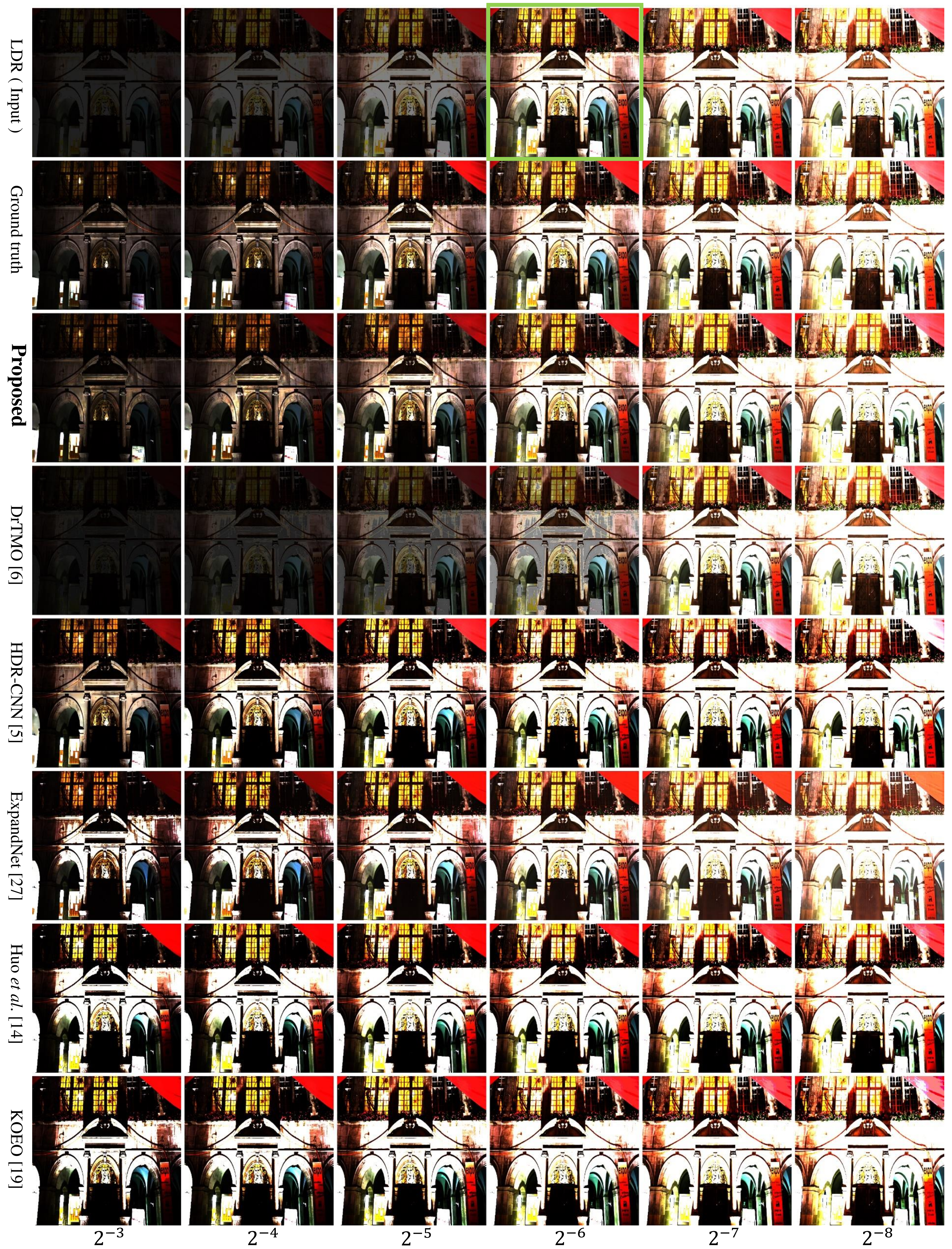}
	\end{center}
    \vspace{-3mm}
    \caption{An example of night views of classic buildings. The proposed method can recover the indoor lighting and the color of the building. }
        \label{fig:Result_o2}
\end{figure*}

\begin{figure*}[t]
	\begin{center}
		\includegraphics[width=16.5cm]{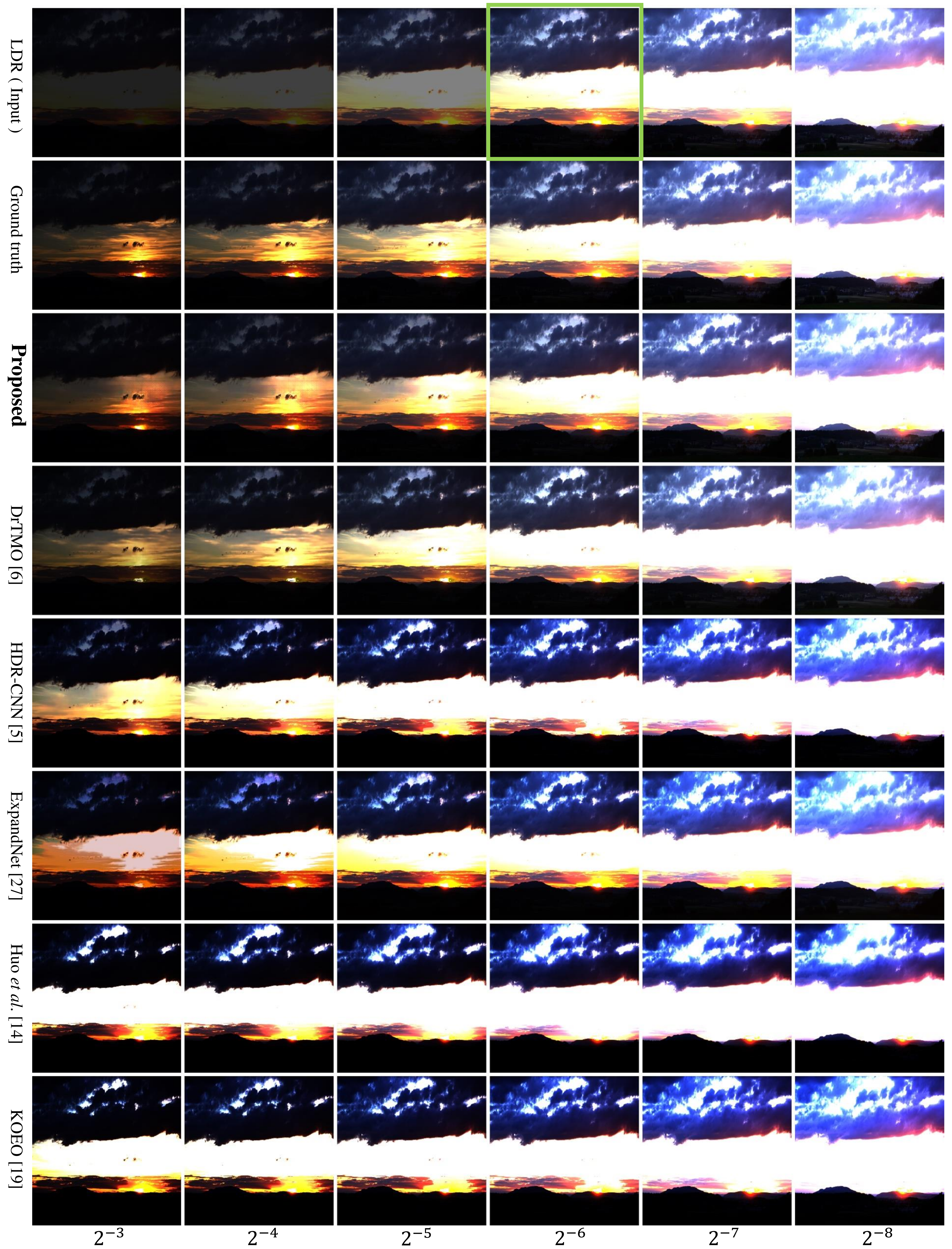}
	\end{center}
    \vspace{-3mm}
    \caption{An example of sunset scenes. The proposed method successfully recovers the texture of the clouds, and the luminance of the sun.}
        \label{fig:Result_o3}
\end{figure*}

\begin{figure*}[t]
	\begin{center}
		\includegraphics[width=16.5cm]{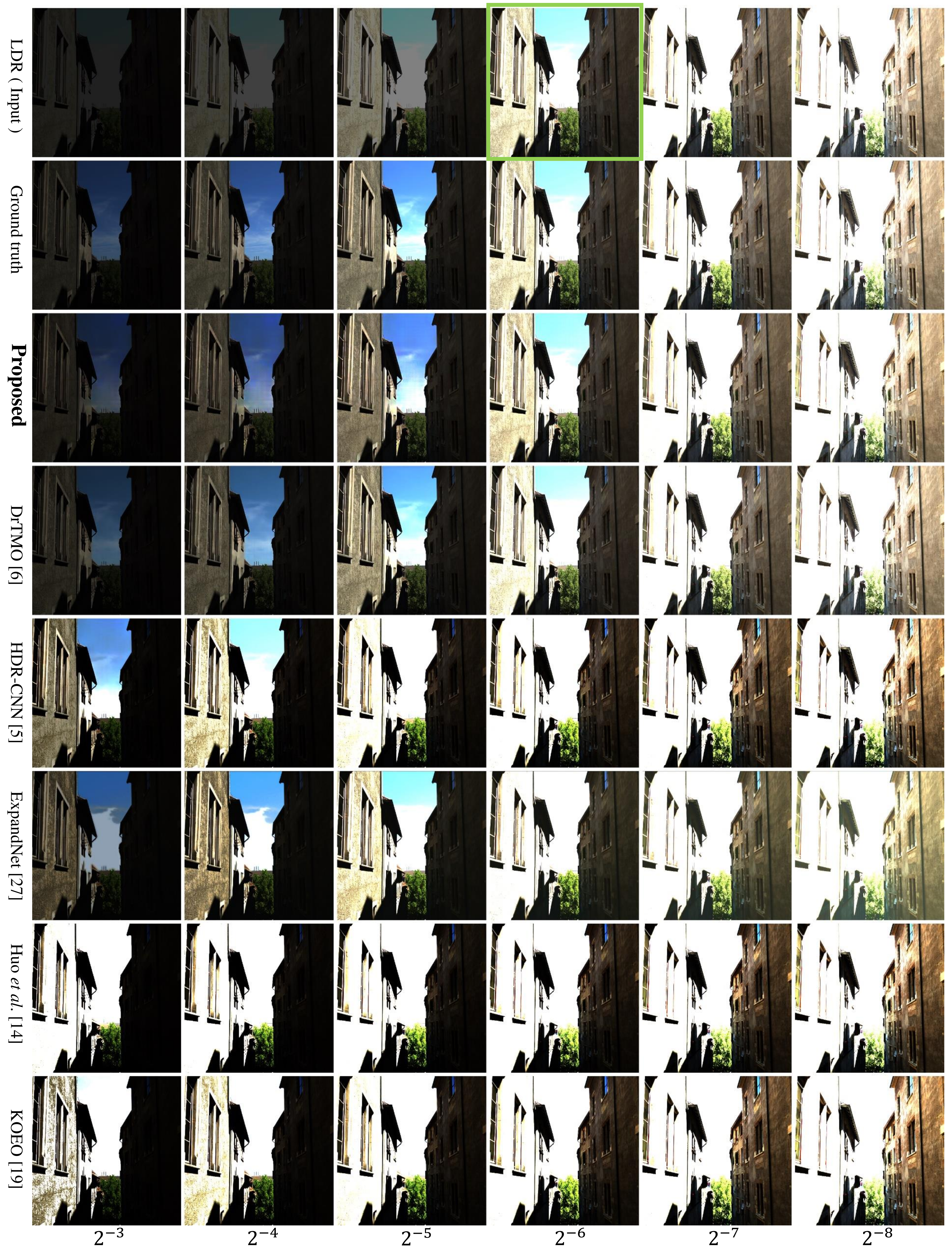}
	\end{center}
    \vspace{-3mm}
    \caption{An example of sunny outdoor scenes. The proposed method can recover the sky, the clouds, and the color of the walls.}
        \label{fig:Result_o4}
\end{figure*}

\begin{figure*}[t]
	\begin{center}
		\includegraphics[width=16.5cm]{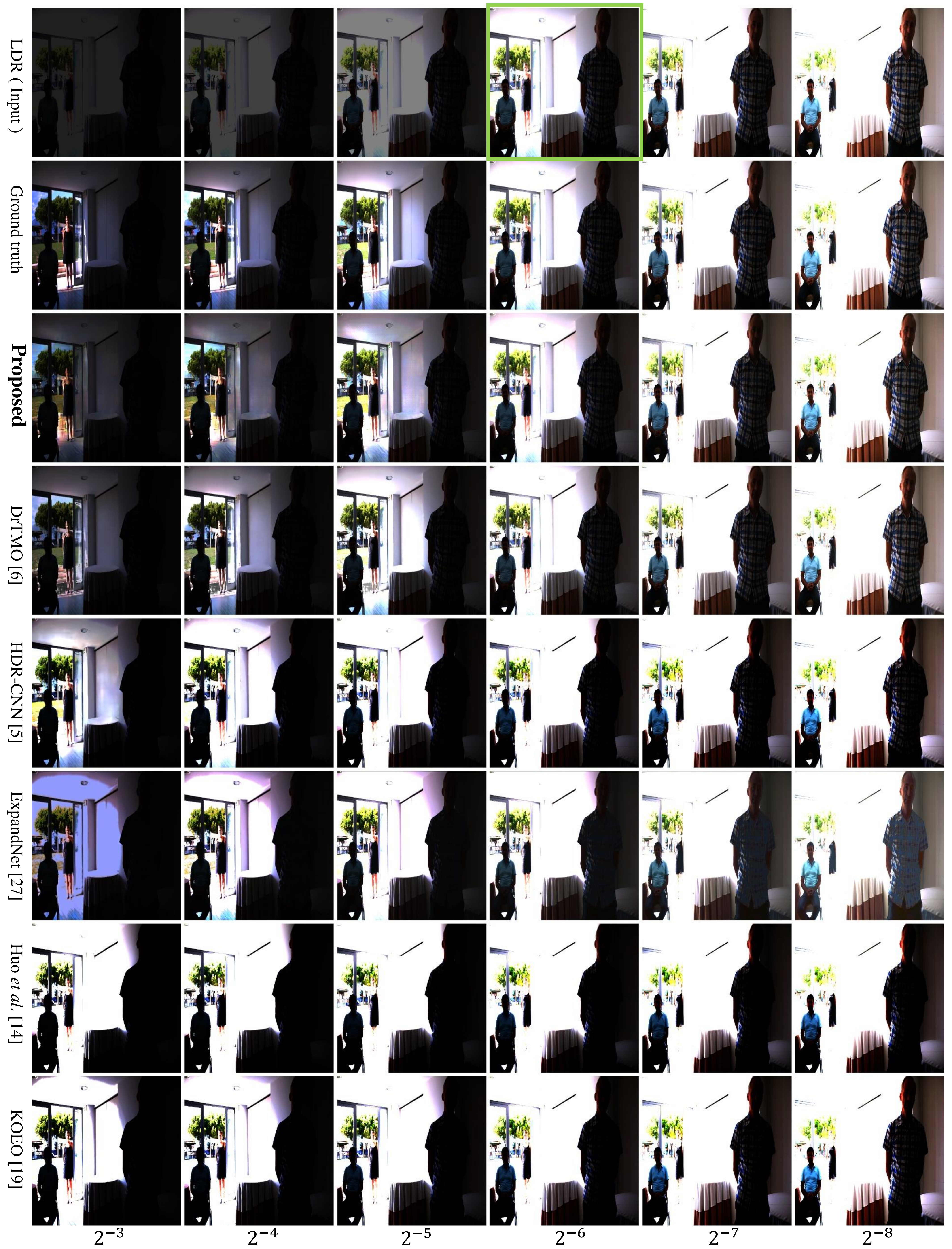}
	\end{center}
    \vspace{-3mm}
    \caption{An example of indoor scenes with windows. The proposed method recovers the intensities of the sky, the trees, the person at the back, and some color of the road.}
        \label{fig:Result_o5}
\end{figure*}

Figures~\ref{fig:Result_o1}--~\ref{fig:Result_o5} are additional visualizations for the range-wise evaluation.
We again compared ours with  the state-of-the-art methods using deep learning, namely DrTMO~\cite{endoSA2017}, HDR-CNN~\cite{HDRsingle2017} and ExpandNet~\cite{Marnerides_2018_EG}, 
and the iTM methods of Huo \etal~\cite{HuoYDB14} and Kovaleski and Oliveira expansion operator (KOEO)~\cite{KovaleskiOliveira2014}. 
As the figures show, the proposed method can reconstruct very bright light intensities that are clearly visible in the range~\footnote{$2^{-3}$ stands for the quantization resolution as explained in Fig.4 of the main text.} of $2^{-3}$, and the recovered intensities are closer to the ground truths in most cases compared to the other methods. 

\subsection{Qualitative evaluation of the proposed method}
\begin{figure*}[t]
	\begin{center}
    \includegraphics[width=16.5cm]{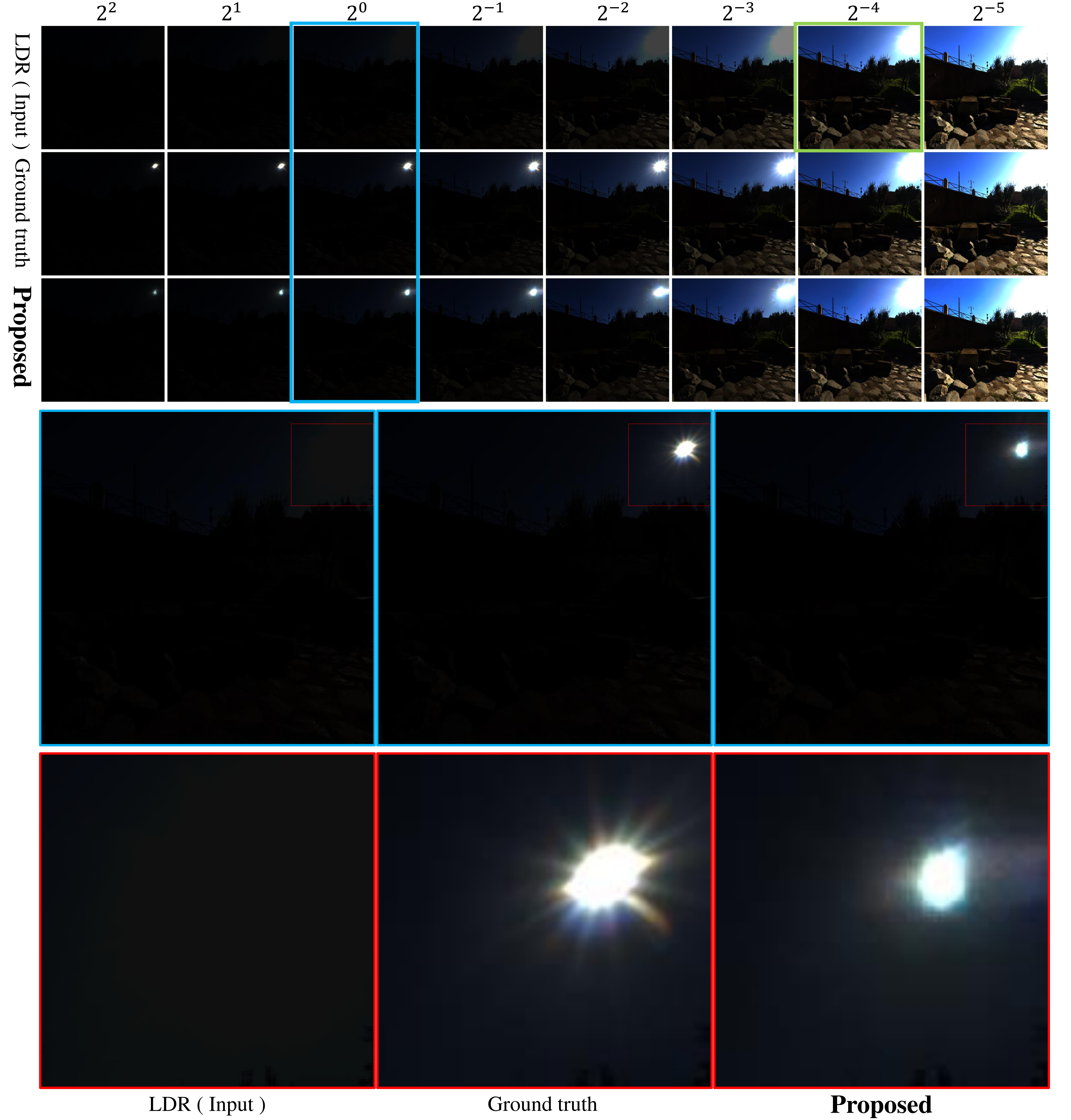}
	\end{center}
    \vspace{-3mm}
    \caption{`Colosseum' from the HDRIhaven dataset. The top row shows multiple exposure levels of an LDR image (the original LDR is highlighted with a green rectangle), where each column corresponds to the levels from $2^{-5}$ to $2^{2}$. The second row shows the ground truth, and the third row shows the HDR images reconstructed by the proposed method. The images in the range of $2^{0}$ are highlighted with a blue rectangle and the fourth row shows the zoomed images of them. The regions highlighted with red rectangles are further zoomed and shown in the bottom row.
The proposed method recovers the light intensities that is still 
observable in the range $64$ times higher than the range of the original image.    
}
        \label{fig:Result_h1}
\end{figure*}

\begin{figure*}[t]
	\begin{center}
   		\includegraphics[width=16.5cm]{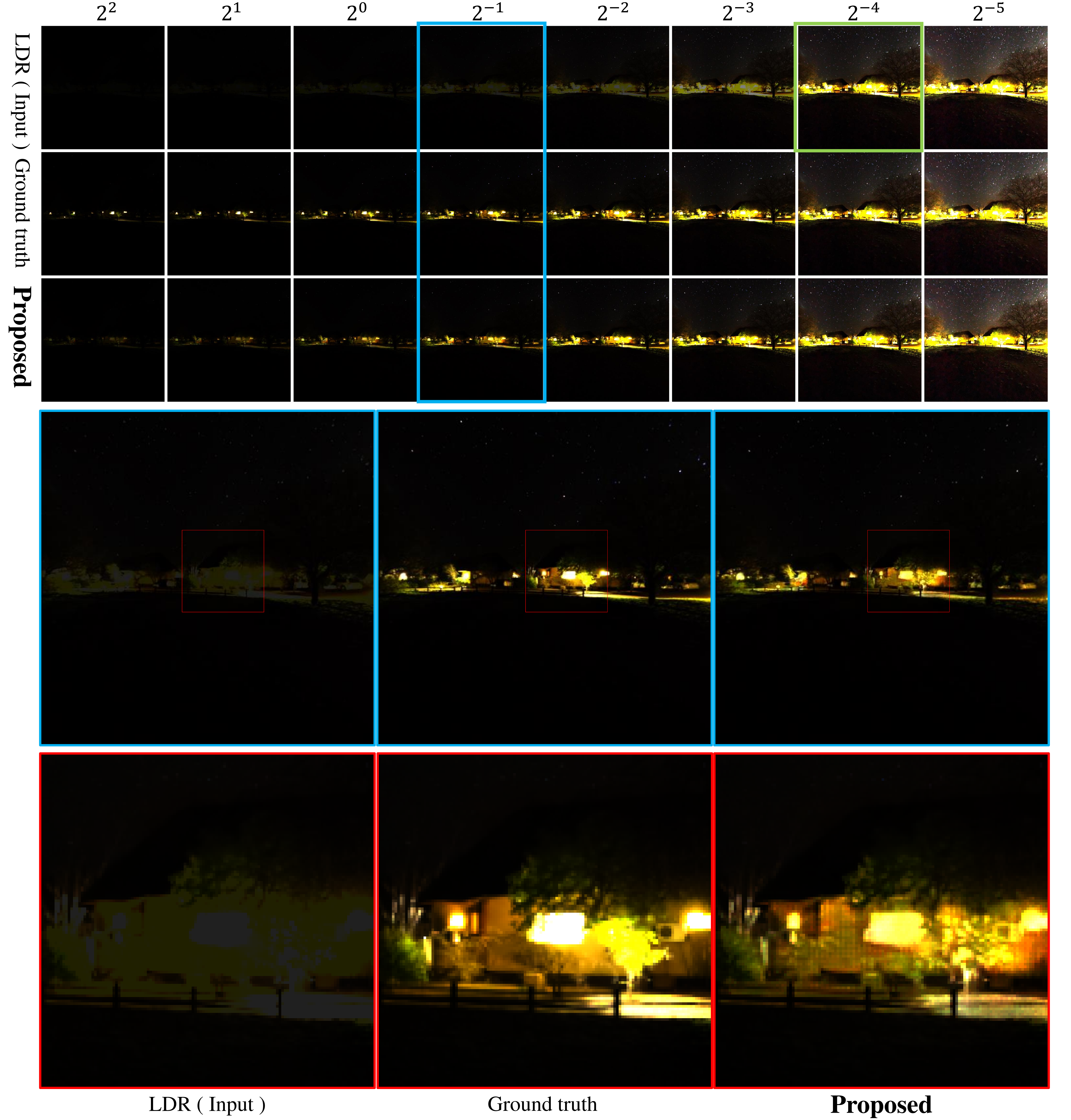}
	\end{center}
    \vspace{-3mm}
    \caption{`Satara night.' The proposed method recovers the intensities of not only the light sources but also non-illuminating objects such as trees and houses around the light sources. }
        \label{fig:Result_h2}
\end{figure*}

\begin{figure*}[t]
	\begin{center}		\includegraphics[width=16.5cm]{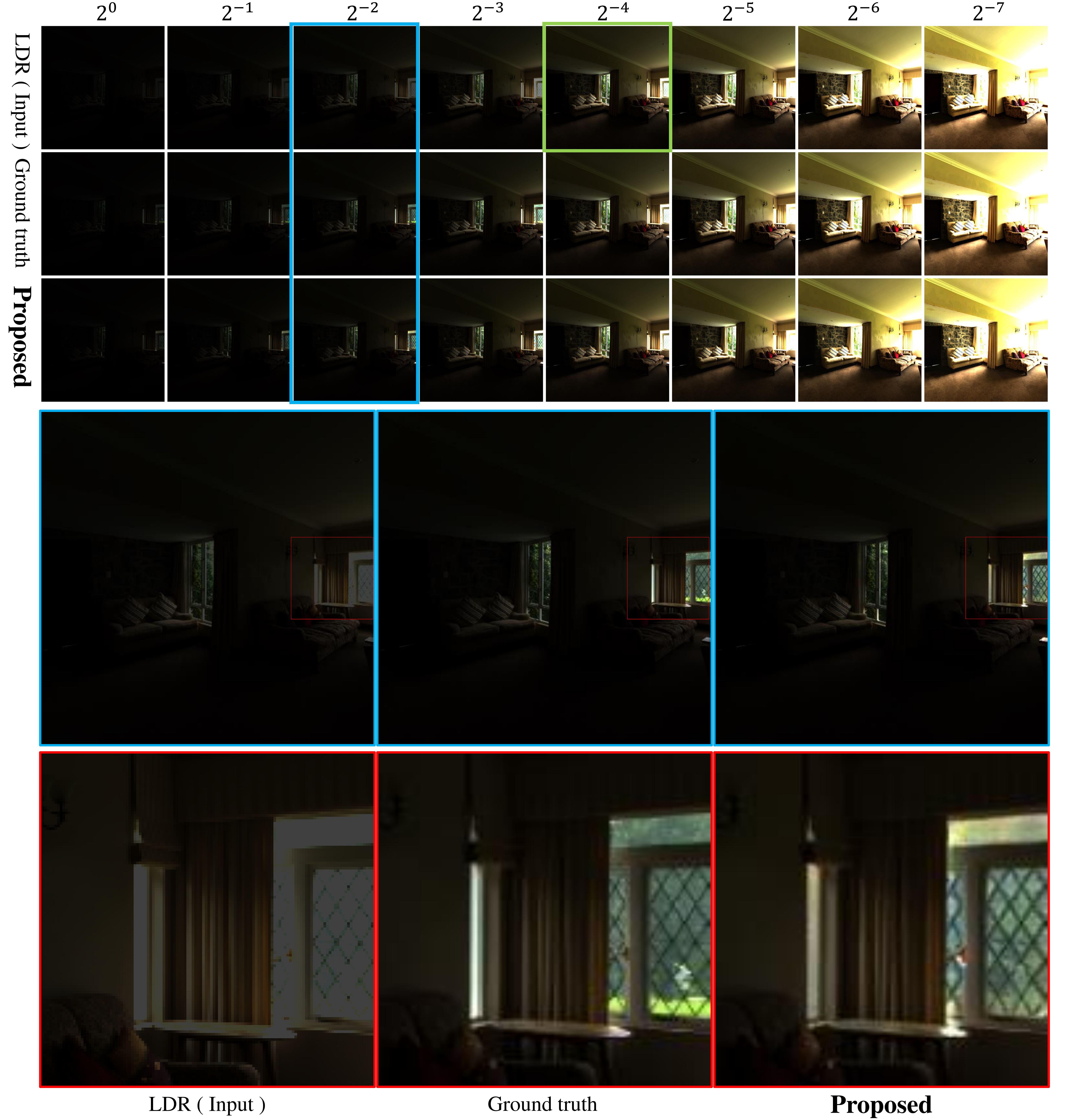}
	\end{center}
    \vspace{-3mm}
    \caption{`Lythwood lounge.' The proposed method inpainted the trees outside the window, which are totally lost in the LDR image. }
        \label{fig:Result_h5}
\end{figure*}

\begin{figure*}[t]
	\begin{center}		\includegraphics[width=16.5cm]{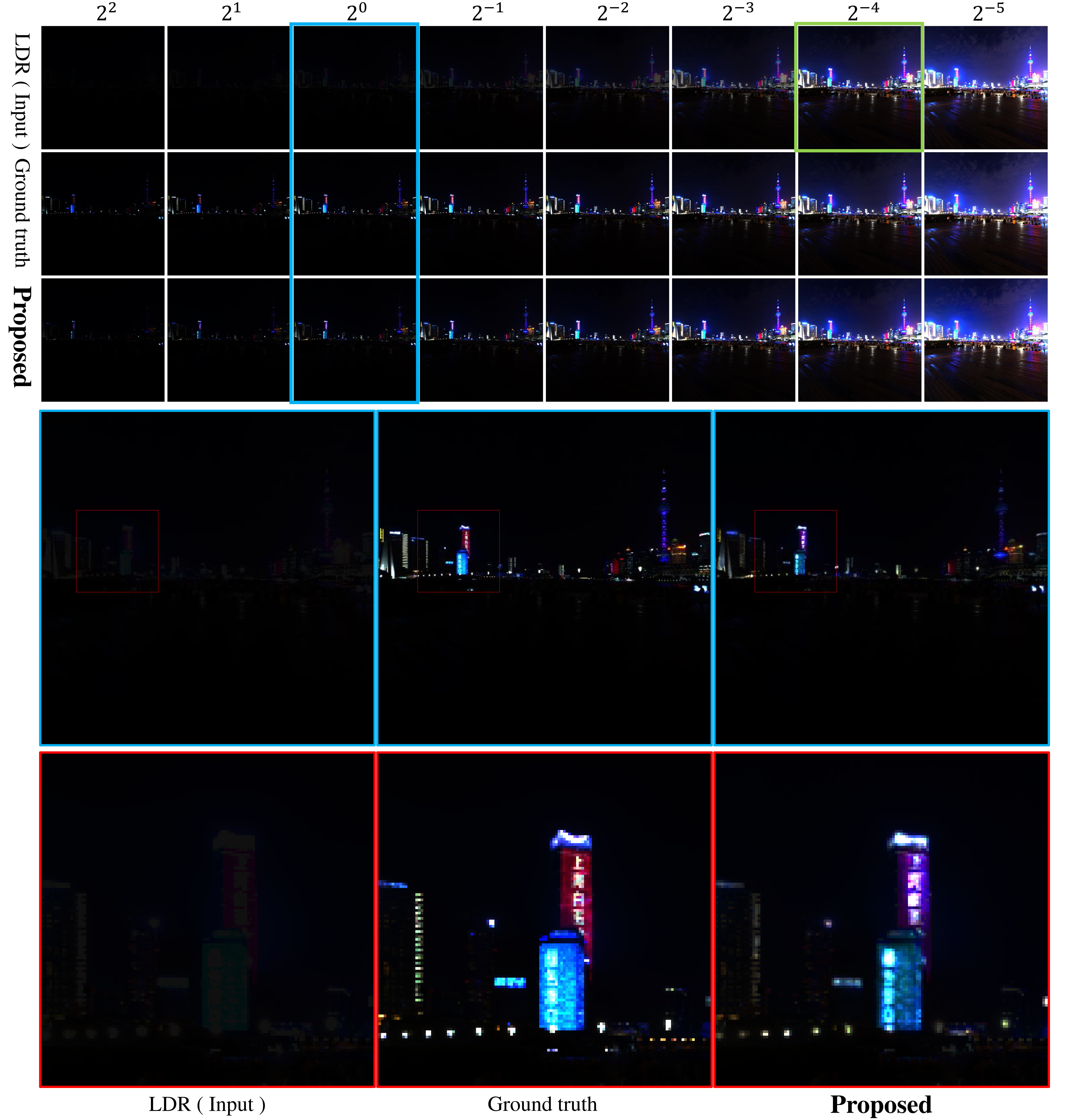}
	\end{center}
    \vspace{-3mm}
    \caption{`Shanghai bund.' The proposed method recovers 
    the buildings' red and blue light-ups. The hue of the red color is slightly shifted, but visually the restoration is natural. }
        \label{fig:Result_h6}
\end{figure*}

We further show qualitative results using images in HDRIhaven~\cite{hdrihaven}, which includes HDR images taken with a wide range of exposures (around 20 EVs). 
Figures~\ref{fig:Result_h1}--~\ref{fig:Result_h6} show the inputs, the ground truths, and the reconstructed images by the proposed method. 
The results show that our method can reconstruct the light intensities well, in a very high dynamic range and with complex textures.

\end{document}